\documentclass[11pt]{article}
\usepackage[margin=0.9in]{geometry}
\usepackage{hyperref}
\usepackage{url}
\usepackage{float}
\usepackage{enumitem}
\usepackage{graphicx}
\usepackage{subcaption}
\usepackage{amsmath}
\usepackage{bm}
\usepackage{natbib}
\usepackage{amsthm, amssymb}

\usepackage{authblk}
\usepackage{caption}
\usepackage{subcaption}

\title{Do Mice Grok? Glimpses of Hidden Progress During Overtraining in Sensory Cortex}

\author{
Tanishq Kumar\textsuperscript{1} \ \ Blake Bordelon\textsuperscript{2,4} \ \ Cengiz Pehlevan\textsuperscript{2,3,4} \ \ 
Venkatesh N. Murthy\textsuperscript{3,4,5} \ \ Samuel J. Gershman\textsuperscript{3,4,6} \\ 
Harvard University
\\
}

\date{} 

\footnotetext[1]{Harvard College; correspondence to \texttt{tkumar@college.harvard.edu}}
\footnotetext[2]{Department of Applied Mathematics, SEAS}
\footnotetext[3]{Center for Brain Science}
\footnotetext[4]{Kempner Institute for the Study of Natural and Artificial Intelligence}
\footnotetext[5]{Department of Molecular and Cellular Biology}
\footnotetext[6]{Department of Psychology}

\begin{document}

\maketitle

\begin{abstract}
Does learning of task-relevant representations stop when behavior stops changing? Motivated by recent theoretical advances in machine learning and the intuitive observation that human experts continue to learn from practice even after mastery, we hypothesize that task-specific representation learning can continue, even when behavior plateaus. In a novel reanalysis of recently published neural data, we find evidence for such learning in posterior piriform cortex of mice following continued training on a task, long after behavior saturates at near-ceiling performance (``overtraining"). This learning is marked by an increase in decoding accuracy from piriform neural populations and improved performance on held-out generalization tests. We demonstrate that class representations in cortex continue to separate during overtraining, so that examples that were incorrectly classified at the beginning of overtraining can abruptly be correctly classified later on, despite no changes in behavior during that time. We hypothesize this hidden yet rich learning takes the form of approximate margin maximization; we validate this and other predictions in the neural data, as well as build and interpret a simple synthetic model that recapitulates these phenomena. We conclude by showing how this model of late-time feature learning implies an explanation for the empirical puzzle of overtraining reversal in animal learning, where task-specific representations are more robust to particular task changes because the learned features can be reused.
\end{abstract}

\newpage 

\section{Introduction}

Understanding how representations evolve during task learning is a fundamental question in both neuroscience and machine learning. While the dynamics of learning are standard objects of study in both fields, the representational dynamics at play long \textit{after} ceiling performance is reached on the training task are much less commonly studied. 

In deep learning, this has changed with the discovery of the grokking phenomenon \citep{power2022grokking}, whereby neural networks first fit their training data to high accuracy, then many epochs of training later  (``overtraining"), abruptly generalize. As a phenomenon that epitomizes how neural networks can behave in sharp and unexpected ways, it has been under intense study \citep{nanda2023progress, liu2022omnigrok, varma2023explaining}. Emerging theories posit that generalization occurs due to delayed feature learning taking place when the network has already achieved ceiling accuracy on the training task \citep{kumar2023grokking,lyu2023dichotomy}. 

In psychology, there exists a long line of work on learning tasks at all levels of expertise, including after task mastery \citep{ericsson1993role, newell1981mechanisms, fitts1967human, mackintosh1969further, richman1972overtraining, mead1973effect}. Many of these works attempt to address the intuitive observation that human experts in a variety of domains continue to learn on tasks at which they have achieved near-ceiling performance. There has also been much work on using deep learning to model perceptual learning tasks, to study the degree to which biological and artificial neural networks exhibit similar behavior on this class of tasks \citep{wenliang2018deep, bakhtiari2019can, yashar2017feature}.

The nature of representational changes at play during overtraining on learned tasks have been tackled at a fine grained (theorems proven about certain classes of deep networks) and coarse grained (qualitative explanations in psychology) levels. Such an understanding, however, is broadly missing at an intermediate level: that of experimental neuroscience. This is because there are several difficulties in naively testing the hypothesis that representations continue to evolve during overtraining in cortex. First, for many animals, recording neurons is invasive and risks damaging the animals, so recordings are typically only taken at the end of training to study the end-time learned representation \citep{yuste2015neuron, kim2016long}. Second, experiments in systems neuroscience do not typically design stimuli as ``training" and ``test" in the way datasets are constructed in deep learning, so making claims about learning and generalization is not possible. We elaborate on further difficulties, as well as the connection of our phenomenon of interest to the related phenomenon of representational drift, in Appendix \ref{appdx: difficulties}. 
    
Here, we revisit the neural data from a recent work with precisely the desirable qualities above: \citep{berners2023experience}, who study neural activity during learned odor discrimination task in mouse piriform cortex. The mice are trained to discriminate one target odor from hundreds of nontarget odors, then continue to be trained for weeks on the same task after reaching behavioral mastery (``overtraining"), with neural firing rate data recorded during this time. After overtraining, they are given held-out test examples. Our starting point in this work are two empirical observations in \citep{berners2023experience} that we seek to model and explain.  

\begin{itemize}
    \item Decoding accuracy of training labels from population activity increases throughout overtraining.
    \item Mice overtrained for longer tend to do better on held-out test examples after the overtraining period is complete.  
\end{itemize}

Our goal in this work is to understand the representational dynamics in cortex at play underlying these observations, and to what extent, if any, they are similar to those at play in deep neural networks where learning continues after training accuracy reaches ceiling \citep{power2022grokking}. Our contributions are the following:

\begin{itemize}
    \item We find that target-nontarget odor class representations in mouse piriform cortex continue actively separating during overtraining while behavior remains unchanged, loosely resembling the ``hidden progress" \citep{barak2022hidden, nanda2023progress} that often taking place in synthetic deep learning settings.  
    \item We find that the margin of the maximum margin classifier at each day of overtraining is increasing in mouse cortex, implying that points (odor trials) sufficiently close to the optimal decision boundary may be classified incorrectly at the beginning of overtraining, but correctly at the end, despite no outward change in behavior. 
    \item We construct a synthetic model of piriform cortex performing this task exhibits these properties, as well as the grokking phenomenology. We interpret the resulting dynamics and use targeted ablations to trace the continued learning to margin maximization.  
    \item We use our insight to suggest a new, fine-grained and neural explanation for the overtraining reversal effect, an empirical puzzle from experimental psychology in which animals overtrained on a task learn its reversal more quickly than those not. 
\end{itemize}

We also note that \citet{berners2023experience} are not the first to observe such effects experimentally: similar effects in sharpening of already-learned odor representations was also observed by \citet{shakhawat2014arc, kadohisa2006separate}, so that the phenomena we seek to model in this work  are likely to be robust at least in the setting of mouse olfaction.

\section{Related Work}
\label{others}

\textbf{Lazy/rich learning and margin maximization.} We cast our ideas in the language of ``lazy" and ``rich" regimes of representation learning, introduced in machine learning \citep{chizat2019lazy, jacot2018neural}. Such characterization of learning regimes has become increasingly popular in computational neuroscience \citep{chizat2019lazy, farrell2023lazy, flesch2021rich, flesch2022orthogonal, ito2023multitask}. In machine learning, lazy learning refers to a trained neural network being well approximated by a kernel method in its initial neural tangent kernel (NTK) and the rich regime is when this approximation breaks down as the network learns task-relevant features and its weights move far from initialization. In classification problems, margin maximization can be one type of rich learning \citep{mohamadi2024you, matyasko2017margin}, where the margin of a classifier is defined as the distance from the decision boundary to the nearest data point. There are some subtleties around defining margin maximization in a setting where a decoder is being retrained on an evolving feature map; we state a precise working definition in Section \ref{margin-defn}. 

\textbf{Grokking.} Much theoretical work has been done to understand the grokking phenomenon, discovered by \citet{power2022grokking}, in which task-specific representation learning occurs long after training accuracy has saturated \citep{nanda2023progress, barak2022hidden, liu2022towards}. Current theories explaining grokking are varied: \citet{liu2022omnigrok} claim it is due to weight norm at initialization, \citet{nanda2023progress} suggest it is driven by weight decay, \citet{varma2023explaining} suggest it is due to competition between memorizing and generalizing neural circuits. A flurry of recent theoretical papers unify these views this as late-time feature learning, where networks begin lazy and then abruptly transition into the rich regime much later during training \citep{kumar2023grokking, lyu2023dichotomy}. Recent empirical evidence has accumulated in support of this perspective \citep{edelman2024evolution, clauw2024information, mohamadi2024you}, where rich learning often takes the form of margin maximization in classification tasks. For instance, \citet{morwani2023feature} and \citet{mohamadi2024you} show that on certain classes of tasks, margin maximization during overtraining provably causes grokking.

\section{Revisiting Neural Data From Overtraining}
\label{section:data-analysis}

\begin{figure}\label{fig:decoding-acc}
    \centering{
        
        \includegraphics[width=0.36\linewidth]{unnamed.pdf}
    }
    \hfill
    {
        \includegraphics[width=0.61\linewidth]{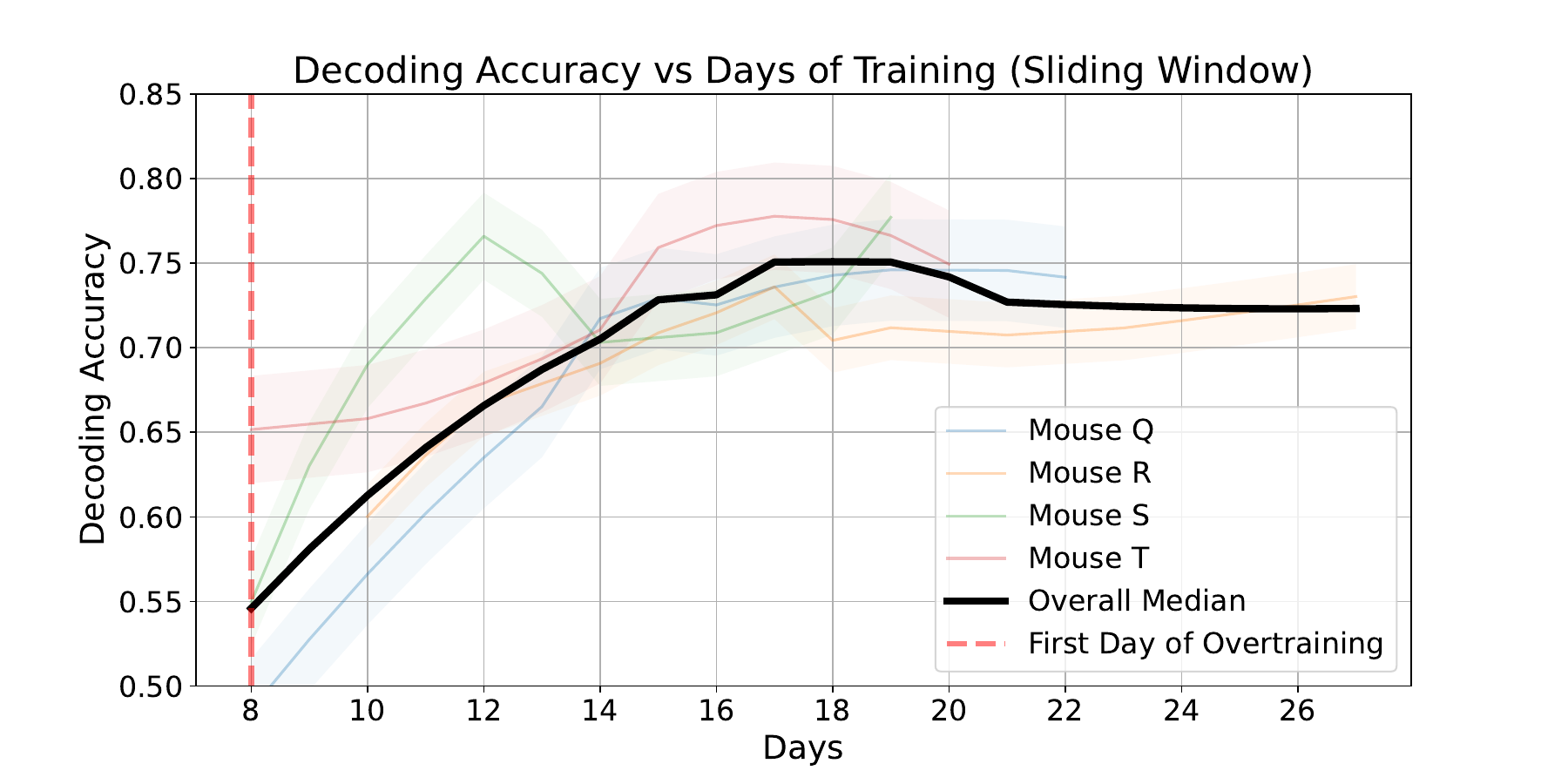}
    }
    \caption{(Left) Behavior of mice on a binary discrimination task of odors, where mice indicate their selected choice by licking left or right. Y-axis is fraction of licks left vs right on a given day. Day 8 is the first day of overtraining. Correct lick for non-target is left, and right for target. The mice can discriminate near-perfectly as overtraining begins. Reproduced with modification from \citep{berners2023experience}. 
    (right) Increase in decoding accuracy from 10-fold linear discriminant analysis for each session. Shaded colored lines in background show standard errors for each mouse. }  
    \label{fig:task-setup-overall}
\end{figure}

\subsection{Setup}

\citet{berners2023experience} collected tetrode recordings from posterior piriform cortex in adult mice during overtraining on a binary odor discrimination task (Figure \ref{fig:task-setup-overall}). The posterior piriform cortex (PPC) is a key brain region involved in olfactory processing \citep{blazing2020odor, srinivasan2017quantitative}. An odor is defined as an $n$-hot vector of length $k$, with $n=3, k = 13$. In other words, a unique combination of three chemicals out of thirteen possible chemicals. The ``target" odor is uniquely defined as consisting of three specific chemicals, where non-targets never contain any of these three. The discrimination task mice are trained on is to distinguish the target from the nontarget odors. After an initial training period of 8 days, the mice master the task, reaching ceiling behavioral performance (see Figure \ref{fig:task-setup-overall}a), yet training continues in the same way for a further 18 days (``overtraining"). This period is when firing rate data is recorded. After this overtraining period is complete, the mice are exposed to held-out odors (``test/probe" examples), which share one odorant with the target odor and therefore are more difficult to correctly classify. 



\subsubsection{Representations of Target/Nontarget Odors Separate During Overtraining}

\begin{figure}
    \centering
    \includegraphics[width=0.9\linewidth]{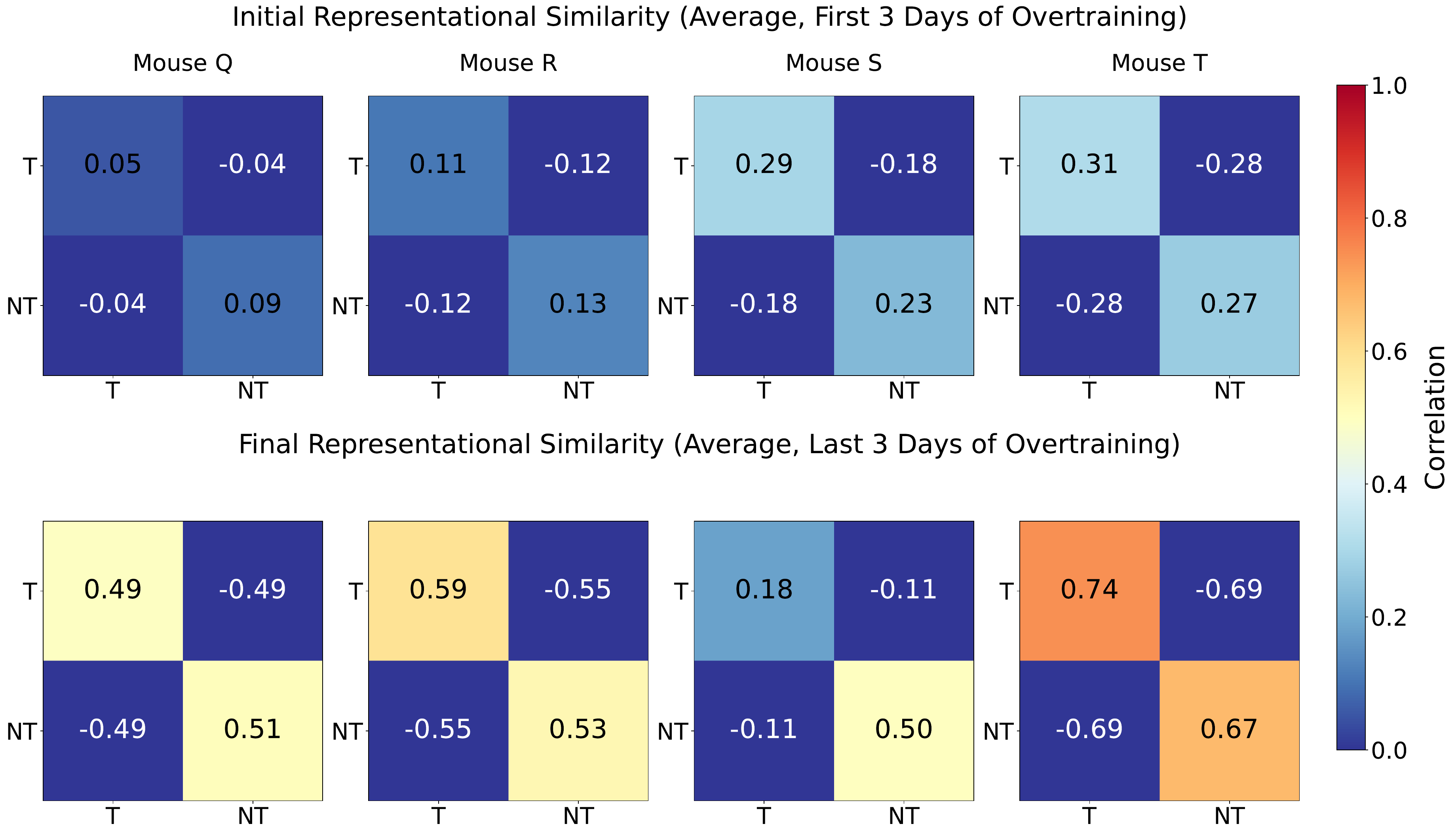}
    \caption{Representational similarity matrices plotting average pairwise correlation between target and nontarget population responses from piriform cortex on the first (top) and last (bottom) days of overtraining. We drop the last few days of data for Mouse T due to inconsistencies in data; methodological details in \ref{appdx: method}. We compare the separations (anticorrelation) above to that of random/ablated baselines in Appendix \ref{appdx: ablations}, finding them significantly larger.}
    \label{fig:rsa}
\end{figure}

 The key object of study throughout the neural data reanalysis will be the population firing rate matrix $X \in \mathbb{R}^{N\times D}$ which contains the neural response vector (activations) for each trial (presented stimulus) on a given-mouse day. There are around $N\approx 200$ trials per day each recording around $D \approx 15$ neurons, and around 15-20 days for each of 4 mice overall. We compute the average dot product (over trials) between the (z-scored) firing rates between target and nontarget trials, finding that representations of target and non-target odors continue to separate over the course of overtraining. We plot the representational similarity matrices at the beginning and end of overtraining for each mouse (Figure \ref{fig:rsa}), with further methodological details deferred to Appendix \ref{appdx: method}. 
 
 The measurement of such representational similarity metrics is standard in neuroscience \citep{kriegeskorte2008representational}, and similar to common interpretability techniques in deep learning that involve probing for directions in activation space \citep{bau2018gan, olah2020zoom, elhage2021mathematical, zhang2023towards} and comparing them, for instance by computing their dot product. The continued separation of class representations between the target and nontarget odors in Figure \ref{fig:rsa} provides evidence that, despite no visible changes in behavior, there is continued representation learning in mouse PPC during overtraining on this task. Such continued learning may aid in generalization on unseen odors, as indeed \citet{berners2023experience} find empirically to be the case.

\begin{figure}
    \centering
    \includegraphics[width=0.85\linewidth]{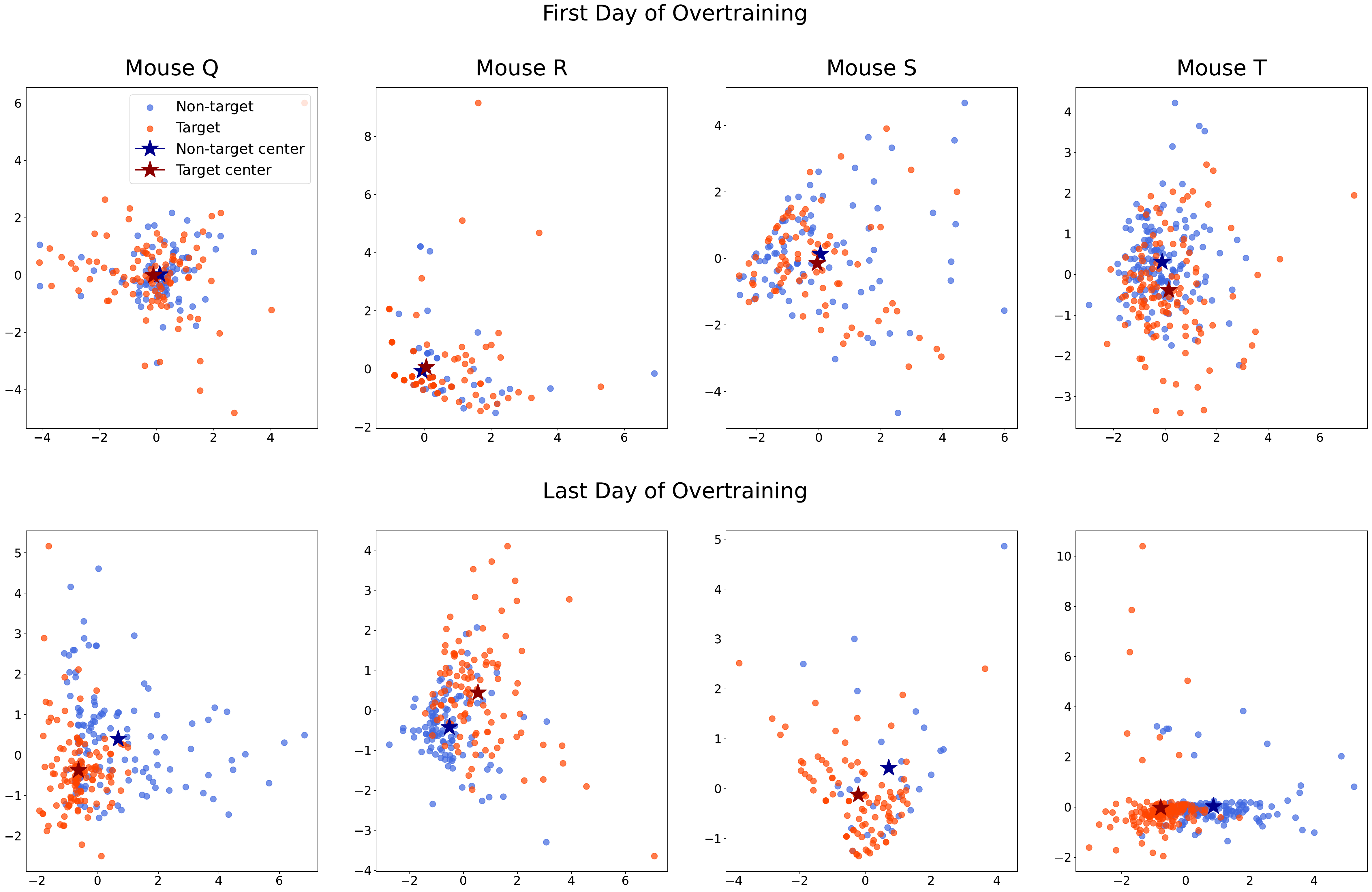}
    \caption{Neural representations (projected onto the top 2 principal components) of target and nontarget odors on the first (top) and last (bottom) day of overtraining for each mouse, showing qualitative separation during overtraining measured quantitatively in the representational similarity matrices in Figure \ref{fig:rsa}. Cluster centers plotted for ease of visual comparison.}
    \label{fig:real-pca}
\end{figure}

What do these task-relevant changes look like visually? In Figure \ref{fig:real-pca}, we plot the projection of odor representations on a given session onto the first 2 principal components of the firing rate matrix for that session, which is trials by neurons for each mouse. We see generally that the bottom row has more separated representations than the top row, providing a qualitative view of how target odors are more distinguishable in PPC representation space after overtraining than before. These results provide a glimpse of the representational dynamics at play as decoder accuracy is increasing, broadly consistent with the increasing separation we see quantitatively in Figure \ref{fig:rsa}. Note that this continued representation learning (``rich feature learning") despite no changes in training accuracy (``behavior") is precisely what characterizes the abrupt generalization found in the grokking phenomenology in deep learning \citep{lyu2019gradient, kumar2023grokking, mohamadi2024you}. 

\label{define-margin}

\subsubsection{Representational Changes Lead to an Improved Maximum Margin Classifier}
\label{margin-defn}

We now take an alternative perspective on the class separation of odor representations and instead examine the robustness of the learned classifier, as measured by its margin, a classical notion of confidence and robustness from statistical learning theory \citep{Bishop2006}.

We begin by clarifying a few distinct notions of margin maximization that apply when a decoder is trained on an evolving hidden representation. 

\begin{enumerate}
    \item A linear decoder is trained on a fixed feature representation $\phi$ for a fixed number of steps, which may or may not involve convergence of the decoder. 
    \item A linear decoder is trained on a fixed feature representation $\phi$ to convergence. 
    \item The feature representation $\phi_t$ \textit{changes} over time, and at \textit{each time step} $t$, a linear decoder is trained on the feature representation to convergence. 
\end{enumerate}

While the margin of the classifier defined by the decoder can increase as the decoder is trained for more steps on a fixed feature representation as in (1), if the margin of the max-margin (converged) classifier is increasing in (3), it implies the features are evolving to allow for a max-margin classifier with larger margin. This latter notion of margin-maximization that involves feature learning is the notion of margin maximization we will use.

We can define this mathematically as follows. Let $\phi_t(x) \in \mathbb{R}^N$ be the representation at time $t$ for data $x$, and let $w(t)$ be the parameters of the max margin SVM solution to the label classification problem with $\phi_t(x_i)$ as input point $i$ and $y_i$ as label. Then the margin $\mathcal M(t)$ takes the form:
\begin{align}
    \mathcal M(t) = \frac{1}{|w(t)|} \ \text{ where } \  w(t) = \min_{w} |w|^2  \ \text{ such that } \   y_i w \cdot \phi(x_i)_t \geq 1.
\end{align}

In these terms, \textit{our hypothesis is that the evolution on odor representations takes the form of (max) margin maximization}.

\begin{figure}
        
    \centering{
    \label{fig:pca-mouse2}
        \includegraphics[width=0.75\linewidth]{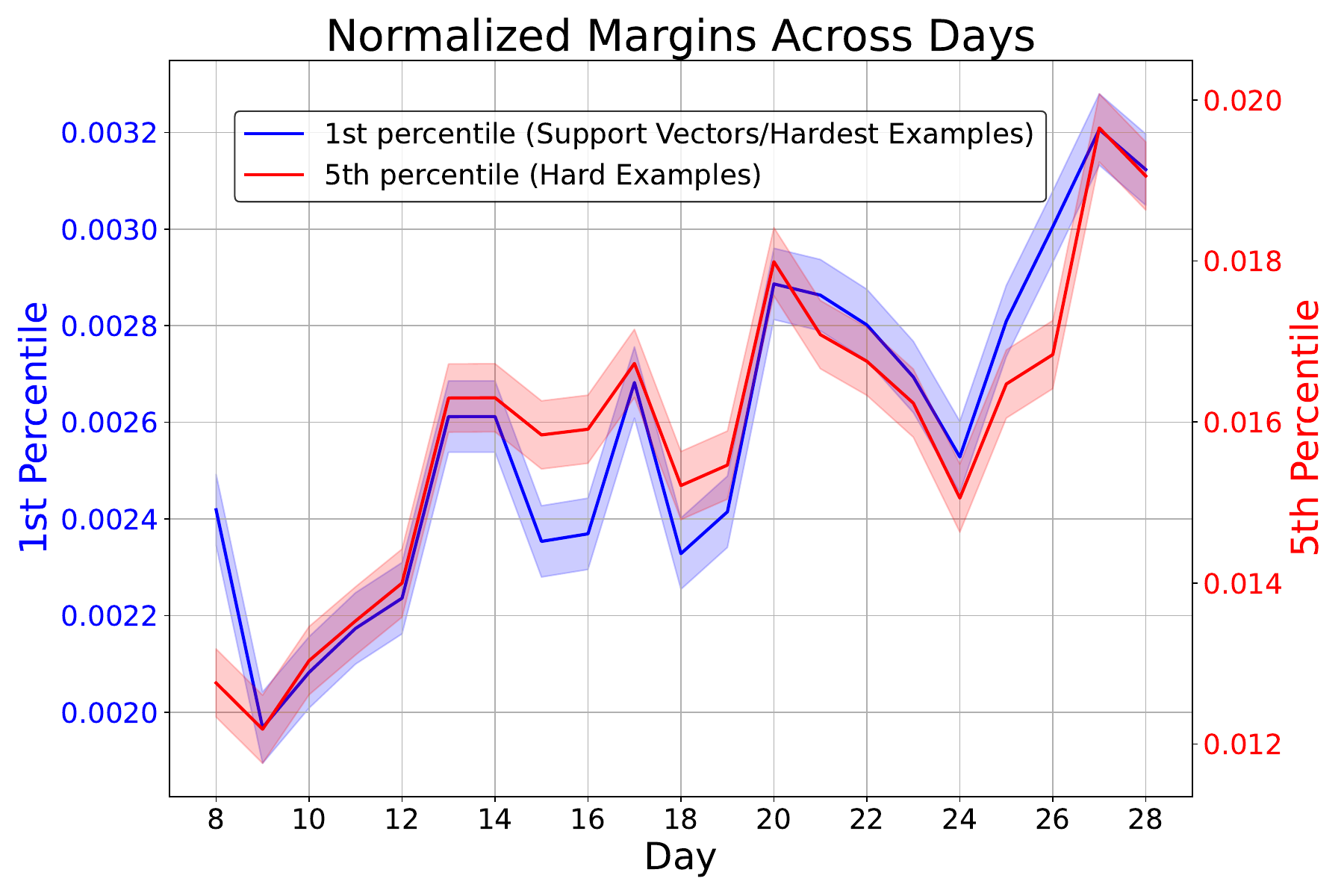}
    }
    \caption{The average distance from the decision boundary within the smallest 1\% and 5\% distances extracted from a linear support vector machine trained on population data from PPC. We normalize margins for each mice so data are comparable across mice, shaded bands show standard errors. }
    \label{fig:margin-mouse}    
\end{figure}

While the ``margin" of a classifier usually refers to the distance to the single closest point, we take it to refer to the distance to the closest 1\% of points, as neural activity underlying individual points (trials) is highly variable. We track the closest points to the learned decision boundary in a linear SVM trained on the same population decoding data to convergence with a fixed number of steps throughout. We measure the average distance from this classifier to the closest 1\% (hardest examples) as well as 5\% (hard examples) of points, finding in Figure \ref{fig:margin-mouse} that these quantities increase substantially during overtraining. 

Therefore, PPC representations evolve to increase the margin of the maximum margin classifier trained on those representations, similar to how many works find that deep neural networks which ``grok" during overtraining on classification tasks are implicitly driven to do so by maximizing margin \citep{morwani2023feature, mohamadi2024you, lyu2019gradient}. This margin-maximization perspective is provocative not only because it provides a geometric and intuitive picture of continued learning in this setting, but because it is suggestive of \textit{why} certain generalizing solutions are learned. For instance, \citet{morwani2023feature, mohamadi2024you, lyu2019gradient} prove in various settings that the reason that neural networks often solve particular tasks (eg. modular arithmetic) using very specific types of features (eg. Fourier features in \citet{nanda2023progress}), is because the regularized optimization trajectories are implicitly biased towards margin-maximizing solutions. 

We emphasize that since test data (unseen test odors) is missing during the training period, we cannot make claims about whether the cortical representation begins lazy or rich (ie. evolve significantly during the initial 7-day learning period compared to initialization), and this is an important avenue for future work and limitation of our reanalysis. While the optimization algorithms at play in brains are poorly understood, it is speculatively possible that some form of ``implicit-bias" towards generalizing solutions may also be the reason that overtrained mice do not just memorize their training odors, but learn representations that are helpful on difficult, held-out examples. 

\section{Theory \& Modeling}

\subsection{A synthetic model of mouse piriform cortex}


\textbf{Setup.} We construct a synthetic version of the odor discrimination task that captures key computational considerations at an abstract level. We consider a more biologically plausible counterpart in Appendix \ref{appdx:biop}, finding the same phenomenology persists in the same way.  We take our earlier definition of odors as $n$-hot vectors of odorants (chemicals) in $\{0, 1\}^k$ that we pass through a random projection, inspired by the functional role of the glomeruli \citep{blazing2020odor}, with $n=10, k=100$. We train a one-hidden layer multilayer perceptron (MLP) on these embeddings to classify the one target odor from 200 non-target odors. We use a ReLU nonlinearity to simulate non-negative firing rates. We also hold out 20 test examples that share an odorant (a single entry in the bit string) with the target. These examples are thus closer in both Hamming and projected space to the target, and therefore harder to discriminate. We train the MLP with vanilla gradient descent on a cross-entropy loss, as is standard in classification. 

\begin{figure}[H]
    \centering
    \includegraphics[width=0.9\linewidth]{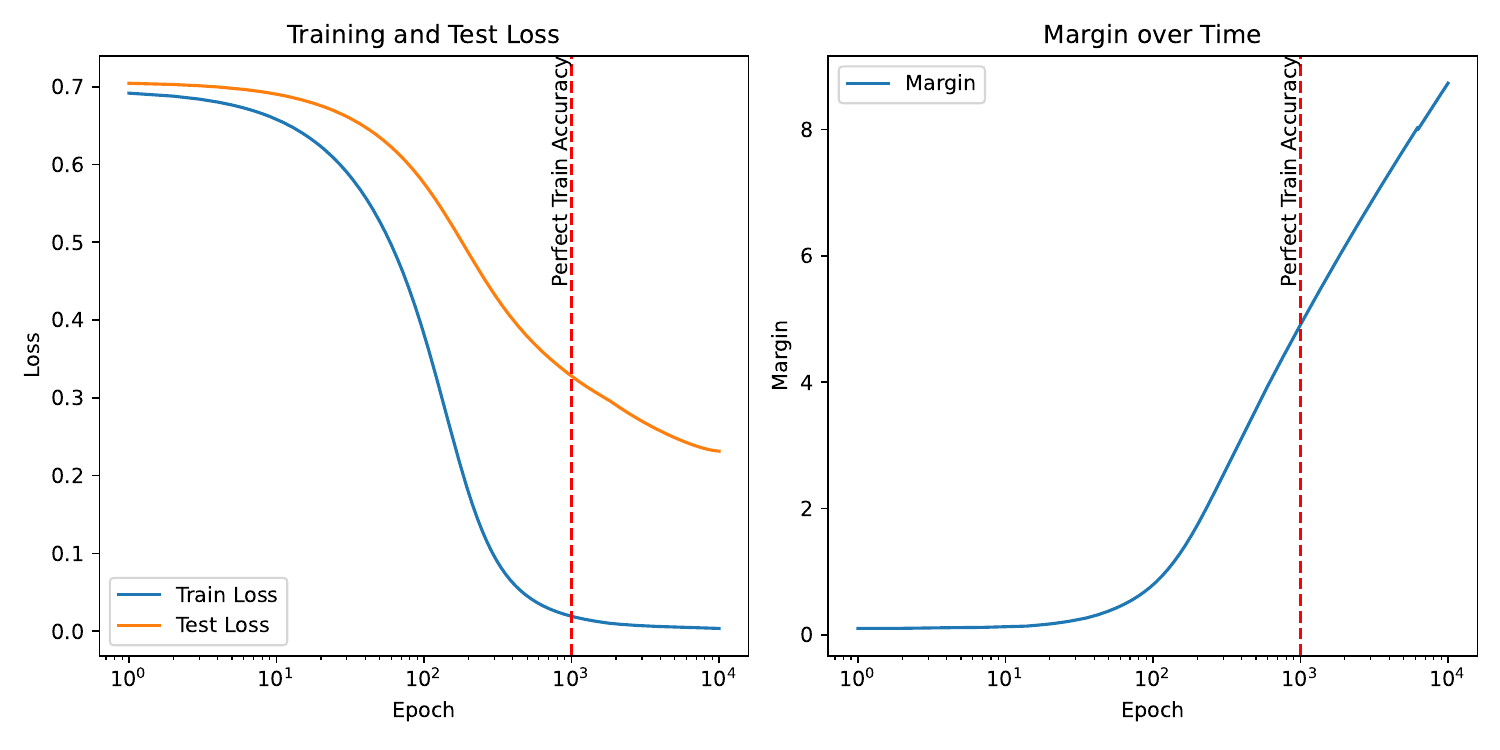}
    \includegraphics[width=0.95\linewidth]{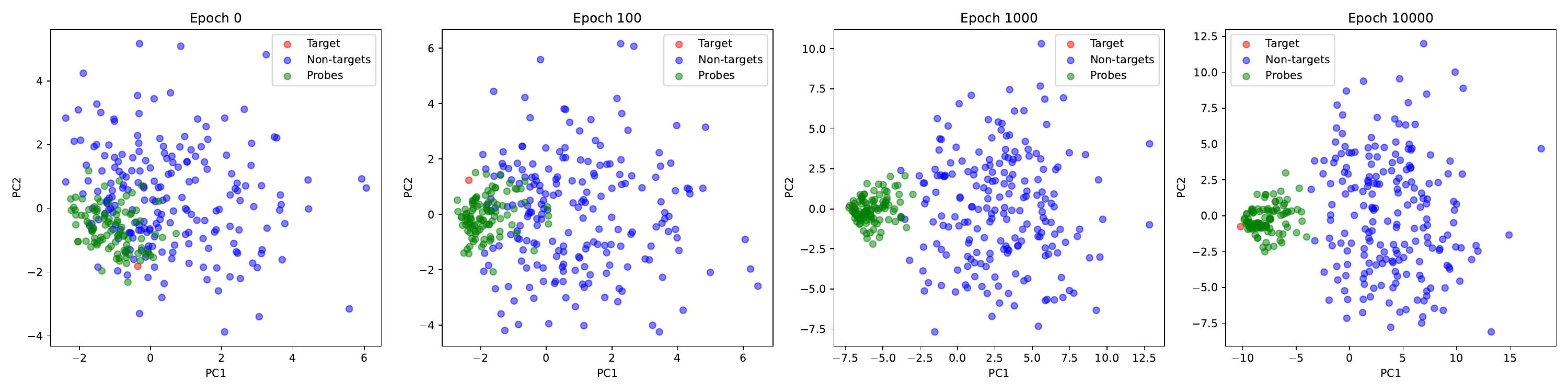}
    \caption{Recapitulating and interpreting mouse piriform cortex dynamics in a simple model. Top left: training and test loss over the course of training. Top right: margin over the course of training. Bottom: projections of the top 2 principal components for several training epochs. Notice how the target and probes (test trials) separate during overtraining (Epochs 1000-9000) despite the target never being trained on a probe example.}
    \label{fig:mlp}
\end{figure}

\begin{figure}
    \centering
    \includegraphics[width=\linewidth]{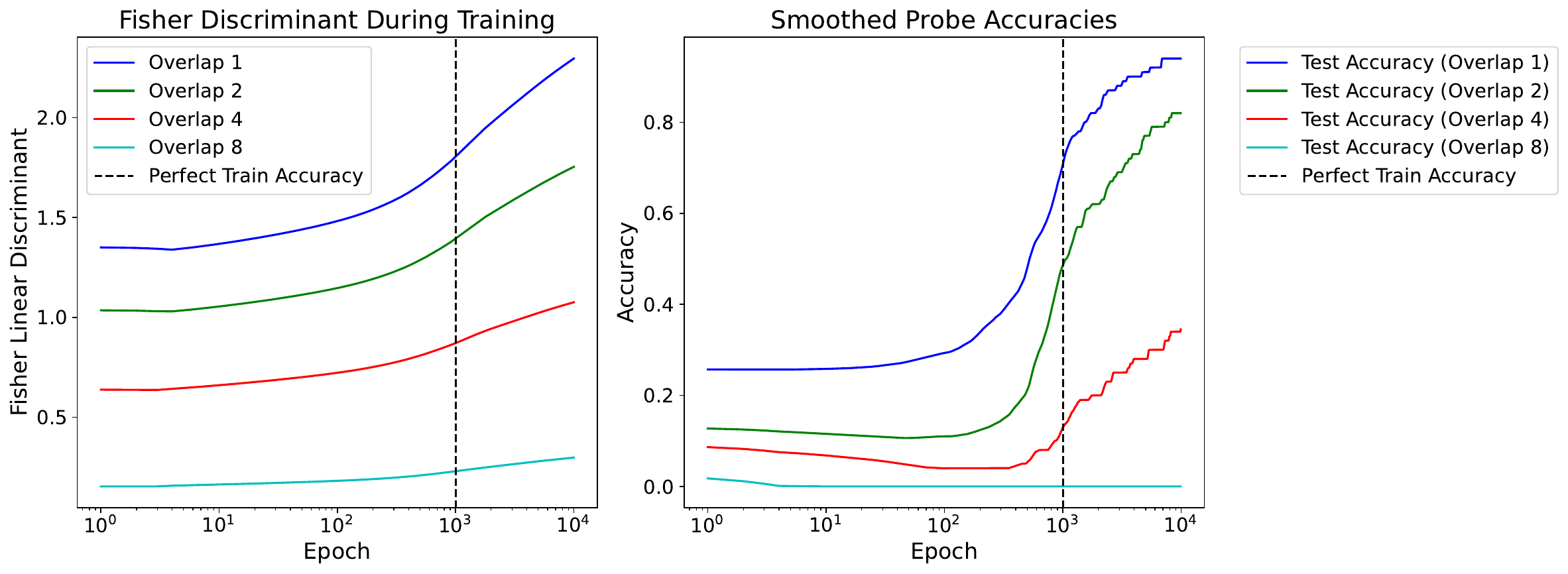}
    \caption{Probe trials are learned in order of increasing overlap with target. (Left) Tracking the Fisher Discriminant (FD) between target and probe classes throughout training. This measures signal to noise in representation space. (Right) The accuracy (``behavioral performance") on increasingly difficult trials over time, exhibiting grokking.
}
    \label{fig:fda-grokking}
\end{figure}

\textbf{Mouse behavior and neural dynamics are captured by an overtrained MLP.} This simple model captures the key computational phenomena at play in piriform data that was the subject of analysis in Section \ref{section:data-analysis}. In Figure \ref{fig:mlp}, we can see that the test loss (on the probe trials) continues to decrease after training loss plateaus at a low value, in line with a continued increase in the classifier margin. The readout layer of our MLP acts as the ``decoder" on the hidden layer representation, which are evolving to allow for an increased max-margin solution to be found. We also see a separation between the target and the nontarget representations emerge during training (up to epoch 1000). However, at epoch 1000, when training loss has converged, the target and probes are not yet easily separable. Strikingly, overtraining on the target-nontarget separation task improves performance on the probe odors, \textit{which are out-of-distribution from the training set}. This happens because the network maximizes margin between the target and nontarget representations during the overtraining period after epoch 1000. This manifests as the red dot (target) gradually separating from the green cluster (test probes) during overtraining. There is an abrupt grokking-like transition in test accuracy (the fraction of probe trials correctly classified) during the overtraining period, Epochs 1000-9000, exactly as we see in Figure \ref{fig:fda-grokking}.

We can test whether margin maximization is the causal factor in this simple model by ablating it and asking whether late-time test loss improvements vanish concomitantly. We do this by replacing the Cross-Entropy objective, which is known to drive late-time margin maximization \citep{lyu2019gradient, lyu2023dichotomy}, with a Hinge loss which has a hard-margin objective instead. This means after the model achieves a margin $C$ on each training point, loss on that point vanishes, so the late-time loss incentive for margin maximization is removed. We find in this ablated setting margin saturates approximately when training loss does, and that test-loss saturates at this point in training as well, so that the grokking-like effect is ablated. Mathematical details and plots are in Appendix \ref{appdx:hard-margin}, supporting the notion of a causal link.

\textbf{Feature learning in the synthetic model.} We increased the dimension of the task ``odor" vector as well as the number of nonzero entries so that we can introduce many types of ``probe" trials that share increasing amounts of odors (overlap) with the target, allowing us to vary difficulty. Note that our model is never trained on the probes, but monitoring probe accuracy allows us to measure how margin maximization on the original nontarget odors (which are guaranteed to have zero overlap with the target) can be helpful in solving more difficult trials. 

In Figure \ref{fig:fda-grokking}, we track the Fisher Discriminant (FD) between target and probe classes, as well as the accuracy on discriminating the two throughout training. The FD is a statistical quantity that uses the linear combination of the features that maximize the separation between two classes, and represents the direction in the feature space along which the projected class means are maximally separated while minimizing the variance within each class. This is the quantity that allows the linear discriminant analysis decoding accuracy to increase; the portion of the lines after low train loss is reached mimics how the median LDA posterior probability on the correct class increases in Figure \ref{fig:task-setup-overall} (right). Mathematical details about the Fisher Discriminant and its relation to margin, LDA, and more are deferred to Appendix \ref{appdx:math}.

\textbf{Possible biological drivers of late time learning.} Certain types of loss functions have been proven in a deep learning setting to drive late-time margin maximization \citep{soudry2017implicit, lyu2019gradient}. Here, we speculate on biologically plausible components of the objective in cortex that may plausibly serve a similar role. A simple but plausible driver of late-time learning is the set of energetic/metabolic constraints on biological organisms. Such constraints been suggested to give rise to a variety of neural phenomena \citep{sterling2015principles}, for example disentangled representations \citep{pehlevan2017blind, Olshausen1996Emergence, Plumbley2003Algorithms, whittington2023disentanglement}. 
Pertinently, such constraints, in the form of weight decay in machine learning, are often the driving force for grokking in some settings \citep{nanda2023progress, liu2022omnigrok}. These constraints force neural networks to deviate from using a lazy (kernel) method, and force the weights to move in task-relevant directions \citep{lyu2019gradient, kumar2023grokking}. Another possibility in \citet{berners2023experience} is a very small supervised error signal from occasional mistakes made by the mice for idiosyncratic reasons that could plausibly be driving late-time learning (See Figure \ref{fig:task-setup-overall}). 

Finally, a key plausible driver of late-time learning in cortex is the existence of unsupervised terms in the implicit objective of organisms that penalize low confidence predictions. For instance, cross-entropy loss is nonzero even when an example is correctly classified because it can be seen as penalizing the \textit{uncertainty} of model predictions, rather than merely \textit{correctness}. The fact that the loss is nonvanishing even when predictions are correct serves as a signal to guide task-relevant learning during overtraining. If biological learning systems store uncertainty in predictions and/or use them for behavior, as indeed is known to be the case in rodent cortex \citep{Kepecs2008} and brains more generally \citep{lak2017midbrain, fetsch2014effects, komura2013responses}, then such implicit penalties on uncertainty will serve a similar functional role to an unsupervised term that keeps loss nontrivial even as training accuracy saturates. This nonvanishing loss during behavioral plateau can plausibly drive continued feature learning during overtraining.

\subsection{Rich Learning During Overtraining can Explain Overtraining Reversal}

Does our model of the dynamics underlying learning during overtraining make any new predictions outside our immediate setup? One concrete prediction of late-time feature learning is faster adaptation under task reversal for any learned task. In machine learning this means fast adaptation of a predictor to distribution shift of the form $y(x) \mapsto -y(x)$. If a network learns features to compute $y(x)$, it is possible many such features can be reused to compute $-y(x)$. This is not true if the network learns $y(x)$, for instance, in the lazy regime, in which case $y$ was fit using the initial features, which were not task-dependent, so cannot be advantageously ``repurposed" for a reversed variant of the task. This particular prediction about reversal learning may seem as oddly specific to a machine learning audience, but in fact overtraining reversal has been a topic of study by experimental pyschologists for decades \citep{mackintosh1969further, richman1972overtraining, mead1973effect}. Theoretical work in psychology offers qualitative cognitive explanations in terms of attention allocation \citep{lovejoy1966analysis}, where our model of overtraining naturally posits a stronger, more fine-grained, neural model for overtraining reversal that is the first of its kind to our knowledge. We begin by defining the phenomenon. 

\label{section:reversal}
\textbf{The Overtraining Reversal Effect.} The reversal effect is the empirical finding that animals that are overtrained on a task more quickly adapt when the task is reversed. Concretely, when animals are given a perceptual discrimination problem and the rewarded mapping is reversed, those animals that were overtrained on the original mapping adapt to the task reversal in fewer trials than those that were not overtrained, see \citep{mackintosh1969further} for further discussion.\footnote{We note that there has been debate on whether the phenomenon is general across species and tasks \citep{beck1966overtraining, warren1978overtraining}.} We propose a simple explanation for this phenomenon in terms of feature learning in the modern machine learning sense. 

\begin{figure}
    \centering
    \begin{subfigure}[b]{0.45\linewidth}
        \includegraphics[width=\linewidth]{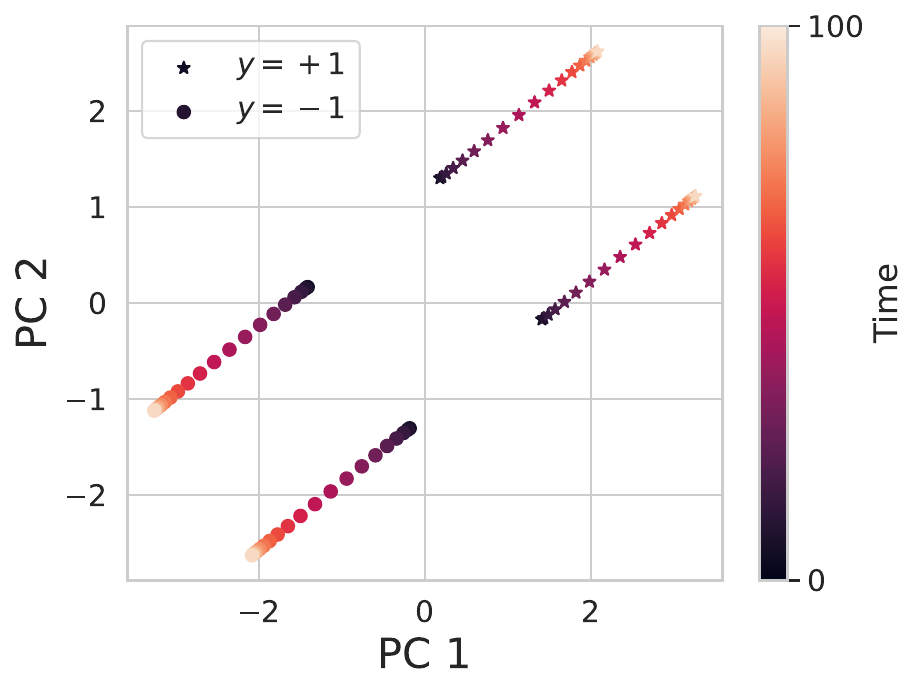}
        \caption{Pretraining Feature Movement ($t<T$)}
    \end{subfigure}
    \hfill
    \begin{subfigure}[b]{0.45\linewidth}
        \includegraphics[width=\linewidth]{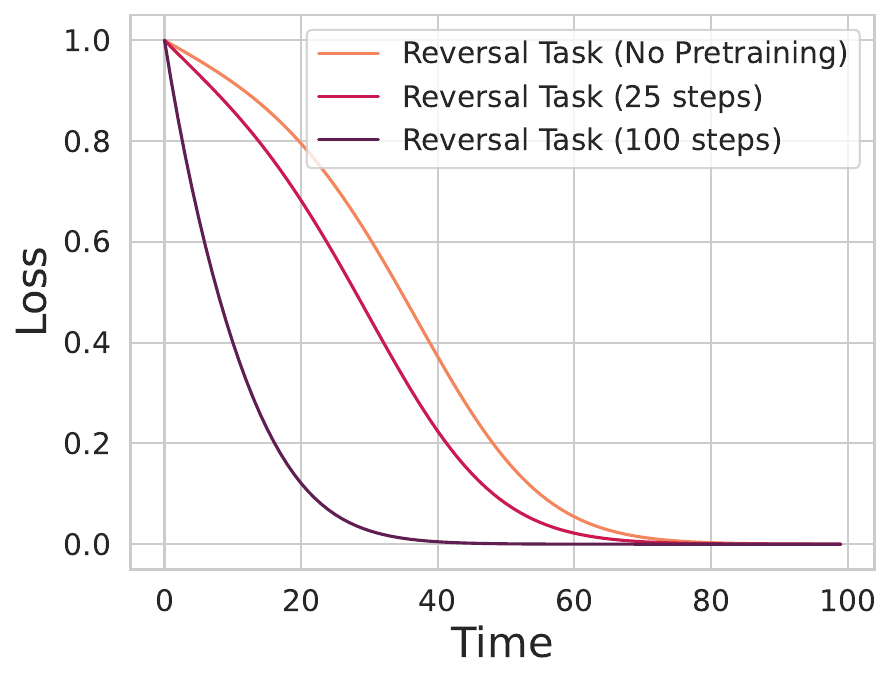}
        \caption{Overtraining = Faster Reversal ($t>T$)}
    \end{subfigure}
    \caption{A simple two-layer model explains faster reversal learning after overtraining on the original task. (a) We plot the original hidden representations (projected from $N$ hidden dimensions to the top two principal components) of a width $N = 250$ network for $4$ odor mixtures, with two odors positive and two odors negatively labeled. (b) When training a new readout on the reversal task $-y(x)$ after pretraining for $t$ steps on the original labels $y(x)$, we find that additional pretraining steps $T$ accelerate learning the reversal task, consistent with theory.}
    \label{fig:reversal}
\end{figure}

\textbf{A Mathematical Model.} We illustrate this with the simplest possible network model that exhibits this effect, a two layer linear network \citep{saxe2013exact}. Reversal learning proceeds in two steps. First, we train a two layer linear network with $N$ hidden neurons $f(\bm x) = \frac{1}{N \gamma_0} \bm w^2 \cdot \bm h(\bm x)$, where $\bm h(\bm x) = \bm W^1 \bm x$ is optimized for $T$ steps on task $y(x)$. After this, a new readout vector $\bm v$ is introduced to give a new output $f_{\text{rev}}(\bm x) = \frac{1}{N \gamma_0} \bm v \cdot \bm h(\bm x)$ which is trained on the reversed task $-y(\bm x)$. The matrix $\bm W^1$ can be updated in the second phase of learning as well. At large width $N$, we can invoke mean field theory ideas \citep[][see Appendix F.1.1]{bordelon2022self}, allowing us to describe the hidden kernel $K(\bm x, \bm x') = \frac{1}{N} \bm h(\bm x) \cdot \bm h(\bm x')$ dynamics in both phases of training. For whitened data and squared error loss, it suffices to compute the projection of the kernel $\bm K$, $K_y = \bm y^\top \bm K \bm y$, and the projection of the predictions $f_y = \bm y^\top \bm f$ along the direction of the labels $\bm y$, which gives:
\begin{align}
       K_y(t) = \sqrt{1 + \gamma_0^2 f_y(t)^2 }  \ , \ \partial_t f_y(t) = 2 \sqrt{1 + \gamma_0^2 f_y(t)^2 } \ (y - f_y(t)).
\end{align}
We show that the kernel evolves only in this $\bm y \bm y^\top$ direction in Figure \ref{fig:reversal} (a), where points separate by their target category. These dynamics are run from time $t = 0$ to time $T$, which gives a terminal value of the kernel alignment $K_y(T)$. In the second phase of learning ($t > T$) where the new output $f_{\text{rev}}$ is fit, the predictions on the reversal task in the $\bm y$ direction evolve as
\begin{align}
    \partial_t f_{\text{rev}, y}(t) = \sqrt{  (K_y(T) + 1)^2 + 4 \gamma_0^2 f_{\text{rev}, y}(t)^2  } \ ( -y - f_{\text{rev}, y}(t)).
\end{align}
We note that larger kernel-target overlap $K_y(T)$ leads to faster training in this second phase $t > T$, as rate of learning on the second task is a direct function of alignment on the first task. We verify this with simulations in Figure \ref{fig:reversal} (b). Since $K_y(T)$ is monotonically increasing with $T$, longer training on the first task will accelerate learning of the reversed task, providing an explanation for overtraining reversal. The key intuition is that adaptation time depends on alignment $K_y=K_{-y}$, where $\bm y^\top \bm K \bm y = (-\bm y^\top) \bm K \bm (-y)$, so overtraining leads to faster adaptation.

\section{Conclusion}\label{section:conclusion}
We find that recent work in mouse olfaction provides evidence for rich sensory learning taking place under the veil of a behavioral plateau in mice. This suggests that difficult tasks that require significant changes in learned representations may benefit from overtraining, even if behavioral performance remains outwardly unchanging. It also offers a glimpse of shared structure between artificial and biological networks, where phenomenologically similar dynamics appear in both cortex and MLPs. 

Our work has a number of limitations. A key limitation of our work is that the neural evidence we draw upon is observational, and new experiments in systems neuroscience are required to validate this hypothesis. We describe in detail a proposed experimental setup in Appendix \ref{appdx:experiment-outline}. Another is that the number of neurons recorded per mouse-day (around 20) is small, so that trends can be noisy (see Appendix \ref{appdx: ablations} for an example). A third limitation is that we consider only olfactory cortex, and only one type of perceptual discrimination task -- whether such late-time learning is a universal property across animals and tasks remains to be seen. We hope our hypotheses drive further work on learning dynamics in cortex during overtraining in animals. Our study raises several new questions: What are the classes of tasks in which such overtraining is beneficial, and what do they have in common? Can deep learning models of cortex hint at the answer? What is the nature of the implicit objective in sensory cortex driving this late-time representation learning? Altogether, our work aims toward a more complete characterization of the behavior of sensory cortex on perceptual discrimination tasks, and to unify the mechanics underlying such behavior with existing theory and observations in deep learning.

\section{Acknowledgements}
Tanishq Kumar thanks Stefano Fusi and Jacob Zavatone-Veth for helpful discussions and comments. Blake Bordelon is supported by a Google PhD Fellowship. Cengiz Pehlevan is supported by NSF grant DMS-2134157, NSF CAREER Award IIS-2239780, and a Sloan Research Fellowship. This work has been made possible in part by a gift from the Chan Zuckerberg Initiative Foundation to establish the Kempner Institute for the Study of Natural and Artificial Intelligence.

\newpage 
\bibliography{mybib}

\begin{thebibliography}{74}
\providecommand{\natexlab}[1]{#1}
\providecommand{\url}[1]{\texttt{#1}}
\expandafter\ifx\csname urlstyle\endcsname\relax
  \providecommand{\doi}[1]{doi: #1}\else
  \providecommand{\doi}{doi: \begingroup \urlstyle{rm}\Url}\fi

\bibitem[Power et~al.(2022)Power, Burda, Edwards, Babuschkin, and Misra]{power2022grokking}
Alethea Power, Yuri Burda, Harri Edwards, Igor Babuschkin, and Vedant Misra.
\newblock Grokking: Generalization beyond overfitting on small algorithmic datasets.
\newblock \emph{arXiv preprint arXiv:2201.02177}, 2022.

\bibitem[Nanda et~al.(2023)Nanda, Chan, Liberum, Smith, and Steinhardt]{nanda2023progress}
Neel Nanda, Lawrence Chan, Tom Liberum, Jess Smith, and Jacob Steinhardt.
\newblock Progress measures for grokking via mechanistic interpretability.
\newblock \emph{arXiv preprint arXiv:2301.05217}, 2023.

\bibitem[Liu et~al.(2022{\natexlab{a}})Liu, Michaud, and Tegmark]{liu2022omnigrok}
Ziming Liu, Eric~J Michaud, and Max Tegmark.
\newblock Omnigrok: Grokking beyond algorithmic data.
\newblock \emph{arXiv preprint arXiv:2210.01117}, 2022{\natexlab{a}}.

\bibitem[Varma et~al.(2023)Varma, Shah, Kenton, Kram{\'a}r, and Kumar]{varma2023explaining}
Vikrant Varma, Rohin Shah, Zachary Kenton, J{\'a}nos Kram{\'a}r, and Ramana Kumar.
\newblock Explaining grokking through circuit efficiency.
\newblock \emph{arXiv preprint arXiv:2309.02390}, 2023.

\bibitem[Kumar et~al.(2023)Kumar, Bordelon, Gershman, and Pehlevan]{kumar2023grokking}
Tanishq Kumar, Blake Bordelon, Samuel~J Gershman, and Cengiz Pehlevan.
\newblock Grokking as the transition from lazy to rich training dynamics.
\newblock \emph{arXiv preprint arXiv:2310.06110}, 2023.

\bibitem[Lyu et~al.(2023)Lyu, Jin, Li, Du, Lee, and Hu]{lyu2023dichotomy}
Kaifeng Lyu, Jikai Jin, Zhiyuan Li, Simon~Shaolei Du, Jason~D Lee, and Wei Hu.
\newblock Dichotomy of early and late phase implicit biases can provably induce grokking.
\newblock In \emph{The Twelfth International Conference on Learning Representations}, 2023.

\bibitem[Ericsson et~al.(1993)Ericsson, Krampe, and Tesch-R{\"o}mer]{ericsson1993role}
K~Anders Ericsson, Ralf~Th Krampe, and Clemens Tesch-R{\"o}mer.
\newblock The role of deliberate practice in the acquisition of expert performance.
\newblock \emph{Psychological review}, 100\penalty0 (3):\penalty0 363--406, 1993.

\bibitem[Newell and Rosenbloom(1981)]{newell1981mechanisms}
Allen Newell and Paul~S Rosenbloom.
\newblock Mechanisms of skill acquisition and the law of practice.
\newblock In John~R Anderson, editor, \emph{Cognitive skills and their acquisition}, pages 1--55. Lawrence Erlbaum Associates, 1981.

\bibitem[Fitts and Posner(1967)]{fitts1967human}
Paul~M Fitts and Michael~I Posner.
\newblock \emph{Human Performance}.
\newblock Brooks/Cole Publishing Company, 1967.

\bibitem[Mackintosh(1969)]{mackintosh1969further}
Nicholas~John Mackintosh.
\newblock Further analysis of the overtraining reversal effect.
\newblock \emph{Journal of Comparative and Physiological Psychology}, 67\penalty0 (2p2):\penalty0 1, 1969.

\bibitem[Richman et~al.(1972)Richman, Knoblock, and Coussens~1]{richman1972overtraining}
Charles~L Richman, Karol Knoblock, and Wayne Coussens~1.
\newblock The overtraining reversal effect in rats: A function of task difficulty.
\newblock \emph{The Quarterly journal of experimental psychology}, 24\penalty0 (3):\penalty0 291--298, 1972.

\bibitem[Mead(1973)]{mead1973effect}
Philip~G Mead.
\newblock The effect of overtraining on reversal shift behavior in rats reinforced with electrical brain stimulation.
\newblock \emph{Physiological Psychology}, 1\penalty0 (4):\penalty0 330--332, 1973.

\bibitem[Wenliang and Seitz(2018)]{wenliang2018deep}
Li~K Wenliang and Aaron~R Seitz.
\newblock Deep neural networks for modeling visual perceptual learning.
\newblock \emph{Journal of Neuroscience}, 38\penalty0 (27):\penalty0 6028--6044, 2018.

\bibitem[Bakhtiari(2019)]{bakhtiari2019can}
Shahab Bakhtiari.
\newblock Can deep learning model perceptual learning?
\newblock \emph{The Journal of Neuroscience}, 39\penalty0 (2):\penalty0 194, 2019.

\bibitem[Yashar and Denison(2017)]{yashar2017feature}
Amit Yashar and Rachel~N Denison.
\newblock Feature reliability determines specificity and transfer of perceptual learning in orientation search.
\newblock \emph{PLoS Computational Biology}, 13\penalty0 (12):\penalty0 e1005882, 2017.

\bibitem[Yuste(2015)]{yuste2015neuron}
Rafael Yuste.
\newblock From the neuron doctrine to neural networks.
\newblock \emph{Nature Reviews Neuroscience}, 16\penalty0 (8):\penalty0 487--497, 2015.

\bibitem[Kim et~al.(2016)Kim, Zhang, Lecoq, Jung, Li, Zeng, and Schnitzer]{kim2016long}
Tae~Hoon Kim, Yajun Zhang, Julien Lecoq, Jason~C Jung, Jun Li, Hongkui Zeng, and Mark~J Schnitzer.
\newblock Long-term optical access to an estimated one million neurons in the live mouse cortex.
\newblock \emph{Cell Reports}, 17\penalty0 (12):\penalty0 3385--3394, 2016.

\bibitem[Berners-Lee et~al.(2023)Berners-Lee, Shtrahman, Grimaud, and Murthy]{berners2023experience}
Alice Berners-Lee, Elizabeth Shtrahman, Julien Grimaud, and Venkatesh~N Murthy.
\newblock Experience-dependent evolution of odor mixture representations in piriform cortex.
\newblock \emph{PLoS Biology}, 21\penalty0 (4):\penalty0 e3002086, 2023.

\bibitem[Barak et~al.(2022)Barak, Edelman, Goel, Kakade, Malach, and Zhang]{barak2022hidden}
Boaz Barak, Benjamin Edelman, Surbhi Goel, Sham Kakade, Eran Malach, and Cyril Zhang.
\newblock Hidden progress in deep learning: Sgd learns parities near the computational limit.
\newblock \emph{Advances in Neural Information Processing Systems}, 35:\penalty0 21750--21764, 2022.

\bibitem[Shakhawat et~al.(2014)Shakhawat, Harley, and Yuan]{shakhawat2014arc}
Amin~MD Shakhawat, Carolyn~W Harley, and Qi~Yuan.
\newblock Arc visualization of odor objects reveals experience-dependent ensemble sharpening, separation, and merging in anterior piriform cortex in adult rat.
\newblock \emph{Journal of Neuroscience}, 34\penalty0 (31):\penalty0 10206--10210, 2014.

\bibitem[Kadohisa and Wilson(2006)]{kadohisa2006separate}
Mikiko Kadohisa and Donald~A Wilson.
\newblock Separate encoding of identity and similarity of complex familiar odors in piriform cortex.
\newblock \emph{Proceedings of the National Academy of Sciences}, 103\penalty0 (41):\penalty0 15206--15211, 2006.

\bibitem[Chizat et~al.(2019)Chizat, Oyallon, and Bach]{chizat2019lazy}
Lenaic Chizat, Edouard Oyallon, and Francis Bach.
\newblock On lazy training in differentiable programming.
\newblock \emph{Advances in neural information processing systems}, 32, 2019.

\bibitem[Jacot et~al.(2018)Jacot, Gabriel, and Hongler]{jacot2018neural}
Arthur Jacot, Franck Gabriel, and Cl{\'e}ment Hongler.
\newblock Neural tangent kernel: Convergence and generalization in neural networks.
\newblock \emph{Advances in neural information processing systems}, 31, 2018.

\bibitem[Farrell et~al.(2023)Farrell, Recanatesi, and Shea-Brown]{farrell2023lazy}
Matthew Farrell, Stefano Recanatesi, and Eric Shea-Brown.
\newblock From lazy to rich to exclusive task representations in neural networks and neural codes.
\newblock \emph{Current opinion in neurobiology}, 83:\penalty0 102780, 2023.

\bibitem[Flesch et~al.(2021)Flesch, Juechems, Dumbalska, Saxe, and Summerfield]{flesch2021rich}
Timo Flesch, Keno Juechems, Tsvetomira Dumbalska, Andrew Saxe, and Christopher Summerfield.
\newblock Rich and lazy learning of task representations in brains and neural networks.
\newblock \emph{BioRxiv}, pages 2021--04, 2021.

\bibitem[Flesch et~al.(2022)Flesch, Juechems, Dumbalska, Saxe, and Summerfield]{flesch2022orthogonal}
Timo Flesch, Keno Juechems, Tsvetomira Dumbalska, Andrew Saxe, and Christopher Summerfield.
\newblock Orthogonal representations for robust context-dependent task performance in brains and neural networks.
\newblock \emph{Neuron}, 110\penalty0 (7):\penalty0 1258--1270, 2022.

\bibitem[Ito and Murray(2023)]{ito2023multitask}
Takuya Ito and John~D Murray.
\newblock Multitask representations in the human cortex transform along a sensory-to-motor hierarchy.
\newblock \emph{Nature neuroscience}, 26\penalty0 (2):\penalty0 306--315, 2023.

\bibitem[Mohamadi et~al.(2024)Mohamadi, Li, Wu, and Sutherland]{mohamadi2024you}
Mohamad~Amin Mohamadi, Zhiyuan Li, Lei Wu, and Danica~J Sutherland.
\newblock Why do you grok? a theoretical analysis of grokking modular addition.
\newblock \emph{arXiv preprint arXiv:2407.12332}, 2024.

\bibitem[Matyasko and Chau(2017)]{matyasko2017margin}
Alexander Matyasko and Lap-Pui Chau.
\newblock Margin maximization for robust classification using deep learning.
\newblock In \emph{2017 International Joint Conference on Neural Networks (IJCNN)}, pages 300--307. IEEE, 2017.

\bibitem[Liu et~al.(2022{\natexlab{b}})Liu, Kitouni, Nolte, Michaud, Tegmark, and Williams]{liu2022towards}
Ziming Liu, Ouail Kitouni, Niklas~S Nolte, Eric Michaud, Max Tegmark, and Mike Williams.
\newblock Towards understanding grokking: An effective theory of representation learning.
\newblock \emph{Advances in Neural Information Processing Systems}, 35:\penalty0 34651--34663, 2022{\natexlab{b}}.

\bibitem[Edelman et~al.(2024)Edelman, Edelman, Goel, Malach, and Tsilivis]{edelman2024evolution}
Benjamin~L Edelman, Ezra Edelman, Surbhi Goel, Eran Malach, and Nikolaos Tsilivis.
\newblock The evolution of statistical induction heads: In-context learning markov chains.
\newblock \emph{arXiv preprint arXiv:2402.11004}, 2024.

\bibitem[Clauw et~al.(2024)Clauw, Marinazzo, and Stramaglia]{clauw2024information}
Kenzo Clauw, Daniele Marinazzo, and Sebastiano Stramaglia.
\newblock Information-theoretic progress measures reveal grokking is an emergent phase transition.
\newblock In \emph{ICML 2024 Workshop on Mechanistic Interpretability}, 2024.

\bibitem[Morwani et~al.(2023)Morwani, Edelman, Oncescu, Zhao, and Kakade]{morwani2023feature}
Depen Morwani, Benjamin~L Edelman, Costin-Andrei Oncescu, Rosie Zhao, and Sham Kakade.
\newblock Feature emergence via margin maximization: case studies in algebraic tasks.
\newblock \emph{arXiv preprint arXiv:2311.07568}, 2023.

\bibitem[Blazing and Franks(2020)]{blazing2020odor}
Robin~M Blazing and Kevin~M Franks.
\newblock Odor coding in piriform cortex: mechanistic insights into distributed coding.
\newblock \emph{Current opinion in neurobiology}, 64:\penalty0 96--102, 2020.

\bibitem[Srinivasan and Stevens(2017)]{srinivasan2017quantitative}
Shyam Srinivasan and Charles~F Stevens.
\newblock A quantitative description of the mouse piriform cortex.
\newblock \emph{bioRxiv}, page 099002, 2017.

\bibitem[Kriegeskorte et~al.(2008)Kriegeskorte, Mur, and Bandettini]{kriegeskorte2008representational}
Nikolaus Kriegeskorte, Marieke Mur, and Peter~A Bandettini.
\newblock Representational similarity analysis-connecting the branches of systems neuroscience.
\newblock \emph{Frontiers in systems neuroscience}, 2:\penalty0 249, 2008.

\bibitem[Bau et~al.(2019)Bau, Zhou, Khosla, Oliva, and Torralba]{bau2018gan}
David Bau, Bolei Zhou, Aditya Khosla, Aude Oliva, and Antonio Torralba.
\newblock Gan dissection: Visualizing and understanding generative adversarial networks.
\newblock \emph{Proceedings of the International Conference on Learning Representations (ICLR)}, 2019.

\bibitem[Olah et~al.(2020)Olah, Cammarata, Schubert, Goh, Petrov, and Carter]{olah2020zoom}
Chris Olah, Nick Cammarata, Ludwig Schubert, Gabriel Goh, Michael Petrov, and Shan Carter.
\newblock Zoom in: An introduction to circuits.
\newblock In \emph{Distill}, 2020.
\newblock https://distill.pub/2020/circuits/.

\bibitem[Elhage et~al.(2021)Elhage, Nanda, Olah, Carter, Wang, and Hernandez]{elhage2021mathematical}
Neil Elhage, A~Nanda, Chris Olah, Shan Carter, Nick Cammarata Gabriel~Schubert Wang, and Curtis Hernandez.
\newblock A mathematical framework for transformer circuits.
\newblock In \emph{Transformer Circuits Thread}, 2021.
\newblock https://transformer-circuits.pub/.

\bibitem[Zhang and Nanda(2023)]{zhang2023towards}
Fred Zhang and Neel Nanda.
\newblock Towards best practices of activation patching in language models: Metrics and methods.
\newblock \emph{arXiv preprint arXiv:2309.16042}, 2023.

\bibitem[Lyu and Li(2019)]{lyu2019gradient}
Kaifeng Lyu and Jian Li.
\newblock Gradient descent maximizes the margin of homogeneous neural networks.
\newblock \emph{arXiv preprint arXiv:1906.05890}, 2019.

\bibitem[Bishop(2006)]{Bishop2006}
Christopher~M. Bishop.
\newblock \emph{Pattern Recognition and Machine Learning}.
\newblock Springer, New York, 2006.
\newblock ISBN 978-0-387-31073-2.

\bibitem[Soudry et~al.(2017)Soudry, Hoffer, and Srebro]{soudry2017implicit}
D~Soudry, E~Hoffer, and N~Srebro.
\newblock The implicit bias of gradient descent on separable data. arxiv.
\newblock \emph{arXiv preprint arXiv:1710.10345}, 2017.

\bibitem[Sterling and Laughlin(2015)]{sterling2015principles}
Peter Sterling and Simon Laughlin.
\newblock \emph{Principles of Neural Design}.
\newblock MIT Press, 2015.

\bibitem[Pehlevan et~al.(2017)Pehlevan, Mohan, and Chklovskii]{pehlevan2017blind}
Cengiz Pehlevan, Sreyas Mohan, and Dmitri~B Chklovskii.
\newblock Blind nonnegative source separation using biological neural networks.
\newblock \emph{Neural computation}, 29\penalty0 (11):\penalty0 2925--2954, 2017.

\bibitem[Olshausen and Field(1996)]{Olshausen1996Emergence}
Bruno~A. Olshausen and David~J. Field.
\newblock Emergence of simple-cell receptive field properties by learning a sparse code for natural images.
\newblock \emph{Nature}, 381\penalty0 (6583):\penalty0 607--609, 1996.

\bibitem[Plumbley(2003)]{Plumbley2003Algorithms}
Mark~D. Plumbley.
\newblock Algorithms for nonnegative independent component analysis.
\newblock \emph{IEEE Transactions on Neural Networks}, 14\penalty0 (3):\penalty0 534--543, 2003.

\bibitem[Whittington et~al.(2023)Whittington, Dorrell, Ganguli, and Behrens]{whittington2023disentanglement}
James~CR Whittington, Will Dorrell, Surya Ganguli, and Timothy Behrens.
\newblock Disentanglement with biological constraints: A theory of functional cell types.
\newblock In \emph{The Eleventh International Conference on Learning Representations}, 2023.

\bibitem[Kepecs et~al.(2008)Kepecs, Uchida, Zariwala, and Mainen]{Kepecs2008}
Adam Kepecs, Naoshige Uchida, Hatim~A Zariwala, and Zachary~F Mainen.
\newblock Neural correlates, computation and behavioural impact of decision confidence.
\newblock \emph{Nature}, 455\penalty0 (7210):\penalty0 227--231, 2008.
\newblock \doi{10.1038/nature07200}.
\newblock URL \url{https://pubmed.ncbi.nlm.nih.gov/18690210/}.

\bibitem[Lak et~al.(2017)Lak, Nomoto, Keramati, Sakagami, and Kepecs]{lak2017midbrain}
Armin Lak, Katsuhiro Nomoto, Mehdi Keramati, Masamichi Sakagami, and Adam Kepecs.
\newblock Midbrain dopamine neurons signal belief in choice accuracy during a perceptual decision.
\newblock \emph{Current Biology}, 27\penalty0 (6):\penalty0 821--832, 2017.

\bibitem[Fetsch et~al.(2014)Fetsch, Kiani, Newsome, and Shadlen]{fetsch2014effects}
Christopher~R Fetsch, Roozbeh Kiani, William~T Newsome, and Michael~N Shadlen.
\newblock Effects of cortical microstimulation on confidence in a perceptual decision.
\newblock \emph{Neuron}, 83\penalty0 (4):\penalty0 797--804, 2014.

\bibitem[Komura et~al.(2013)Komura, Nikkuni, Hirashima, Uetake, and Miyamoto]{komura2013responses}
Yusuke Komura, Akifumi Nikkuni, Naofumi Hirashima, Takuya Uetake, and Atsushi Miyamoto.
\newblock Responses of pulvinar neurons reflect a subject's confidence in visual categorization.
\newblock \emph{Nature Neuroscience}, 16\penalty0 (6):\penalty0 749--755, 2013.

\bibitem[Lovejoy(1966)]{lovejoy1966analysis}
Elijah Lovejoy.
\newblock Analysis of the overlearning reversal effect.
\newblock \emph{Psychological Review}, 73\penalty0 (1):\penalty0 87, 1966.

\bibitem[Beck et~al.(1966)Beck, Warren, and Sterner]{beck1966overtraining}
C.~H. Beck, J.~M. Warren, and R.~Sterner.
\newblock Overtraining and reversal learning by cats and rhesus monkeys.
\newblock \emph{Journal of Comparative and Physiological Psychology}, 62\penalty0 (2):\penalty0 332--335, 1966.
\newblock \doi{10.1037/h0023674}.

\bibitem[Warren(1978)]{warren1978overtraining}
J.~M. Warren.
\newblock Overtraining and extradimensional shift learning by cats.
\newblock \emph{Bulletin of the Psychonomic Society}, 12\penalty0 (3):\penalty0 177--178, 1978.
\newblock \doi{10.3758/BF03329663}.

\bibitem[Saxe et~al.(2013)Saxe, McClelland, and Ganguli]{saxe2013exact}
Andrew~M Saxe, James~L McClelland, and Surya Ganguli.
\newblock Exact solutions to the nonlinear dynamics of learning in deep linear neural networks.
\newblock \emph{arXiv preprint arXiv:1312.6120}, 2013.

\bibitem[Bordelon and Pehlevan(2022)]{bordelon2022self}
Blake Bordelon and Cengiz Pehlevan.
\newblock Self-consistent dynamical field theory of kernel evolution in wide neural networks.
\newblock \emph{Advances in Neural Information Processing Systems}, 35:\penalty0 32240--32256, 2022.

\bibitem[Goltstein et~al.(2021)Goltstein, Reinert, Bonhoeffer, and H{\"u}bener]{goltstein2021mouse}
Pieter~M Goltstein, Sandra Reinert, Tobias Bonhoeffer, and Mark H{\"u}bener.
\newblock Mouse visual cortex areas represent perceptual and semantic features of learned visual categories.
\newblock \emph{Nature neuroscience}, 24\penalty0 (10):\penalty0 1441--1451, 2021.

\bibitem[Rust and DiCarlo(2010)]{rust2010selectivity}
Nicole~C. Rust and James~J. DiCarlo.
\newblock Selectivity and tolerance ("invariance") both increase as visual information propagates from cortical area v4 to it.
\newblock \emph{Journal of Neuroscience}, 30\penalty0 (39):\penalty0 12978--12995, 2010.

\bibitem[Freedman et~al.(2003)Freedman, Riesenhuber, Poggio, and Miller]{freedman2003comparison}
David~J. Freedman, Maximilian Riesenhuber, Tomaso Poggio, and Earl~K. Miller.
\newblock Comparison of categorization of stimuli in the monkey prefrontal cortex: Frontal eye field and inferior convexity.
\newblock \emph{Journal of Neurophysiology}, 90\penalty0 (2):\penalty0 670--682, 2003.

\bibitem[Wang et~al.(2020)Wang, Boboila, Chin, Higashi-Howard, Shamash, Wu, Stein, Abbott, and Axel]{wang2020transient}
Peter~Y Wang, Cristian Boboila, Matthew Chin, Alexandra Higashi-Howard, Philip Shamash, Zheng Wu, Nicole~P Stein, LF~Abbott, and Richard Axel.
\newblock Transient and persistent representations of odor value in prefrontal cortex.
\newblock \emph{Neuron}, 108\penalty0 (1):\penalty0 209--224, 2020.

\bibitem[Barnes et~al.(1997)Barnes, Suster, Shen, and McNaughton]{barnes1997multistability}
Carol~A Barnes, Matthew~S Suster, Jiemin Shen, and Bruce~L McNaughton.
\newblock Multistability of cognitive maps in the hippocampus of old rats.
\newblock \emph{Nature}, 388\penalty0 (6639):\penalty0 272--275, 1997.

\bibitem[Thompson and Best(1990)]{thompson1990long}
LT~Thompson and PJ~Best.
\newblock Long-term stability of the place-field activity of single units recorded from the dorsal hippocampus of freely behaving rats.
\newblock \emph{Brain research}, 509\penalty0 (2):\penalty0 299--308, 1990.

\bibitem[Kentros et~al.(2004)Kentros, Agnihotri, Streater, Hawkins, and Kandel]{kentros2004increased}
Clifford~G Kentros, Naveen~T Agnihotri, Samantha Streater, Robert~D Hawkins, and Eric~R Kandel.
\newblock Increased attention to spatial context increases both place field stability and spatial memory.
\newblock \emph{Neuron}, 42\penalty0 (2):\penalty0 283--295, 2004.

\bibitem[Mankin et~al.(2012)Mankin, Sparks, Slayyeh, Sutherland, Leutgeb, and Leutgeb]{mankin2012neuronal}
Emily~A Mankin, Fraser~T Sparks, Begum Slayyeh, Robert~J Sutherland, Stefan Leutgeb, and Jill~K Leutgeb.
\newblock Neuronal code for extended time in the hippocampus.
\newblock \emph{Proceedings of the National Academy of Sciences}, 109\penalty0 (47):\penalty0 19462--19467, 2012.

\bibitem[Ziv et~al.(2013)Ziv, Burns, Cocker, Hamel, Ghosh, Kitch, Gamal, and Schnitzer]{ziv2013long}
Yaniv Ziv, Laurie~D Burns, Eric~D Cocker, Elizabeth~O Hamel, Kunal~K Ghosh, Lacey~J Kitch, Abbas~El Gamal, and Mark~J Schnitzer.
\newblock Long-term dynamics of ca1 hippocampal place codes.
\newblock \emph{Nature neuroscience}, 16\penalty0 (3):\penalty0 264--266, 2013.

\bibitem[Rule et~al.(2019)Rule, O’Leary, and Harvey]{rule2019causes}
Michael~E Rule, Timothy O’Leary, and Christopher~D Harvey.
\newblock Causes and consequences of representational drift.
\newblock \emph{Current opinion in neurobiology}, 58:\penalty0 141--147, 2019.

\bibitem[Schoonover et~al.(2021)Schoonover, Ohashi, Axel, and Fink]{schoonover2021representational}
Carl~E Schoonover, Sarah~N Ohashi, Richard Axel, and Andrew~JP Fink.
\newblock Representational drift in primary olfactory cortex.
\newblock \emph{Nature}, 594\penalty0 (7864):\penalty0 541--546, 2021.

\bibitem[Qin et~al.(2023)Qin, Farashahi, Lipshutz, Sengupta, Chklovskii, and Pehlevan]{qin2023coordinated}
Shanshan Qin, Shiva Farashahi, David Lipshutz, Anirvan~M Sengupta, Dmitri~B Chklovskii, and Cengiz Pehlevan.
\newblock Coordinated drift of receptive fields in hebbian/anti-hebbian network models during noisy representation learning.
\newblock \emph{Nature Neuroscience}, 26\penalty0 (2):\penalty0 339--349, 2023.

\bibitem[Nakamura et~al.(2021)Nakamura, Gu, Kato, et~al.]{nakamura2021repdrift}
K~Nakamura, Y~Gu, D~Kato, et~al.
\newblock Representational drift in the mouse visual cortex.
\newblock \emph{Nature Communications}, 12:\penalty0 5186, 2021.

\bibitem[Buesing et~al.(2011)Buesing, Bill, Nessler, and Maass]{buesing2011neural}
Lars Buesing, Johannes Bill, Bernhard Nessler, and Wolfgang Maass.
\newblock Neural dynamics as sampling: a model for stochastic computation in recurrent networks of spiking neurons.
\newblock \emph{PLoS computational biology}, 7\penalty0 (11):\penalty0 e1002211, 2011.

\bibitem[Stern et~al.(2018)Stern, Bolding, Abbott, and Franks]{stern2018transformation}
Merav Stern, Kevin~A Bolding, LF~Abbott, and Kevin~M Franks.
\newblock A transformation from temporal to ensemble coding in a model of piriform cortex.
\newblock \emph{Elife}, 7:\penalty0 e34831, 2018.

\bibitem[Krishnamurthy et~al.(2022)Krishnamurthy, Hermundstad, Mora, Walczak, and Balasubramanian]{krishnamurthy2022disorder}
Kamesh Krishnamurthy, Ann~M Hermundstad, Thierry Mora, Aleksandra~M Walczak, and Vijay Balasubramanian.
\newblock Disorder and the neural representation of complex odors.
\newblock \emph{Frontiers in Computational Neuroscience}, 16:\penalty0 917786, 2022.

\bibitem[Mathis et~al.(2016)Mathis, Rokni, Kapoor, Bethge, and Murthy]{mathis2016reading}
Alexander Mathis, Dan Rokni, Vikrant Kapoor, Matthias Bethge, and Venkatesh~N Murthy.
\newblock Reading out olfactory receptors: feedforward circuits detect odors in mixtures without demixing.
\newblock \emph{Neuron}, 91\penalty0 (5):\penalty0 1110--1123, 2016.

\end{thebibliography}
\bibliographystyle{unsrtnat}

\newpage 
\appendix
\section{Appendix}
\subsection{Methodological and implementation details}
\label{appdx: method}

The raw data consists of around 70 sessions, each representing a recording of ~20 random neurons on a given mouse-day. The data is spikes over time for each neuron on each trial, of which there are 200-300 in a session. After odor onset, mice are given 2.5s to lick in response. We count the number of spikes in the final 500ms part of this window and decode on this number, which gives us a $N \times T$ neurons by trials spike count matrix to decode on for each mouse-day. We drop neurons with low (less than 4) number of spikes across all trials classifying them as unresponsive to the odors, we z-score this matrix and use 10-fold cross validation to get decoding accuracy, over 20 iterations. Since session to session data is very noisy, we plot smoothed statistics using a sliding window and plot standard error of all statistics. It should be noted the last few days of mouse $T$ overtrainining are dropped because of notable inconsistencies in neural data such as the decoding accuracy plummeting to far below chance while behavior remained high, suggesting an unknown and idiosyncratic cause for unreliable decoding accuracy in this mouse.\footnote{\citep{berners2023experience} include these unreliable decoding trials in their analysis.} Note that the activity matrix was z-scored across trials, so every neuron contributed equally in our analysis.

\subsection{Testable Predictions \& Experimental Proposals}
\label{appdx:experiment-outline}

\subsubsection{A Concrete Proposal for Piriform Cortex}

While reanalysis of existing data suggests that rich learning during overtraining is plausible, reanalysis on its own insufficient on their own to confidently establish a new empirical phenomenology. In particular, since the focus of \citep{berners2023experience} is studying an increase in target-selectivity and how training history affects the form of the learned representations, targeted experiments attempting to falsify our hypothesis may be helpful. We outline what one such experiment in piriform cortex may look like. Note that such an experimental setup can be applied to any part of sensory cortex in which at least hundreds of neurons can be recorded over several (10+) contiguous days of overtraining. 

We propose an experiment of a similar flavor to that of \citep{berners2023experience}, with several important changes. These changes include: 
\begin{itemize}
    \item Using several, as opposed to one, target odors would prevent the mice from memorizing the target and thus would make the task require more learning of odor identity, leaving more space for rich learning during overtraining.
    \item We would include a spectrum of difficult, held-out, examples, of $n-1$ types if an odorant is an $n$-hot vector of chemicals. Then, we would define $j$-probe trial as those that would share $j$ elements with the target, allowing us to modulate distance in both Hamming and projected space, and therefore difficulty of classification as measured by distance to boundary.
    \item Out of every 200 trials on a given mouse-day, $n-1$ would be probe trials, one for each type. This would be included during the initial training period of 8 days, as well as during and after overtraining, which is not done in \citep{berners2023experience}. This would ensure the test set has minimal impact on learning compared to the training set (which would be $200/n$ times more frequent). 
    \item We would also ensure mice have as much time to recover between overtraining sessions as learning sessions, so the same amount of rich learning can potentially take place, and that conditions during overtraining are kept as similar to those seen during learning, as possible, to compare feature learning on equal footing.  
\end{itemize}

\subsubsection{Important Experimental Details for General Sensory Discrimination Tasks}

In seeking to test our hypothesis, there are a few key experimental details that apply broadly to experiments on sensory cortex beyond olfaction. 

\begin{itemize}
    \item The task needs to involve diverse and novel stimuli. This can be measured by ensuring the ability to decode stimuli from population activity is not significantly higher than change before learning. 
    \item The task needs to be difficult, but possible to master. This is a tricky requirement: here, mice reach up to 90\% behavioral performance, which is close to the upper bound possible given noise and fluctuating conditions. Conversely, the otherwise relevant work in vision of \citep{goltstein2021mouse} trains mice to discriminate gratings based on frequency and orientation. This task is not fully learnable, as the mice reach ceiling of 70\%. Since a primary part of our hypothesis is that this late-time rich learning can persist when task performance is near-ceiling and thus in the absence of large error signal, an error rate of 30\% is too high to test our hypothesis. 
    \item It is crucial to separate a part of the data that is dedicated toward learning the task, namely the ``training" set, and a part of the data dedicated toward testing generalization, namely the ``test" set. As in the piriform setting, constructing some test examples to be particularly challenging by construction is likely to elucidate whether useful yet hidden representation learning takes place in sensory cortex during overtraining. Test trials should be infrequent so their effect on learning is small, but should be included throughout learning so that notions of training and test performance can be determined for the entire training-overtraining-test timeline. 
    \item Chronic recordings should include the evolution of population activity both during training and overtraining. In this work, we cannot examine whether representations begin lazy or rich during learning, since data and probe trials are only available during the overtraining, and not the training, period. We note that mice are good organisms for such a task because larger animals, like macaques, take much longer and are more expensive to train, and chronic and invasive recordings during training can run the risk of causing serious damage to the organism. Conversely, smaller organisms like fruit flies cannot be trained to solve difficult perceptual discrimination tasks with in-vivo neural recordings, in any meaningful sense. 
\end{itemize}

On such setups, our theory predicts continued increases in decoding accuracy for category label for several days after training performance saturates and stops changing noticeably, and separation between task-relevant class representations. 

\subsection{Continued Introduction}
\label{appdx: difficulties}

A third difficulty, in addition to the two mentioned in the main introduction, is that to see genuine representation learning during overtraining, the task must be difficult, but possible to master. In the deep learning setting of grokking, for instance, this is made precise by specifying that grokking occurs under a certain regime of task-model alignment \citep{kumar2023grokking}. Many settings in experimental neuroscience either study passive exposure (no perceptual task learning) \citep{rust2010selectivity, freedman2003comparison}, use tasks which are easy (high initial decoding accuracy) for sensory cortex \citep{wang2020transient}, or use tasks that are too hard in the sense that the animal does only slightly above chance by the end \citep{goltstein2021mouse}. 

Closely related, but distinct from our desired phenomenology of study is that of representational drift. In neuroscience, our understanding of the stability of neural network representations over time has also been revised. Classical work in neuroscience operated under the implicit assumption that neural representations were supported by stable neuronal activity \citep{barnes1997multistability, thompson1990long}. While many neurons do show stable responses to stimuli, the recent discovery of ``representational drift" shows that the population activity underlying some task-relevant stimulus can change over time \citep{kentros2004increased, mankin2012neuronal, ziv2013long, rule2019causes, schoonover2021representational}. Representational drift is commonly conceived as a diffusive process \citep{qin2023coordinated}, and there are arguments that such drift can offer advantages in cortical processing, or confer robustness \citep{nakamura2021repdrift, buesing2011neural}. Our focus is instead on the \textit{learning} of representations during overtraining on a task, and therefore our phenomenology of interest is orthogonal to, but may potentially co-occur with, representational drift.

\subsection{Mathematical Details on Fisher Discriminant, LDA, and Margin}
\label{appdx:math}

Here we provide some background on the key statistical quantities we track, both in piriform cortex and our synthetic model, and their relations to each other. The Fisher Linear Discriminant (FD) is a statistical quantity that measures the ratio of cluster center separation to intra-cluster variance. Informally, it quantifies the signal to noise (SNR) in a binary classification problem in representation space, and is a quantity that Linear Discriminant Analysis (LDA) relies on for classification. It is particularly useful when the goal is to reduce the dimensionality of data while preserving as much class-discriminative information as possible.

Consider a dataset consisting of \( n \) samples, each of which belongs to one of two classes: \( C_1 \) and \( C_2 \). Let \( x_i \in \mathbb{R}^d \) denote the \( d \)-dimensional feature vector of the \( i \)-th sample. The FD seeks a projection vector \(w \in \mathbb{R}^d \) such that when the data is projected onto this vector, the separation between the projected means of the two classes is maximized relative to the projected variance within each class.

Formally, the projection of a data point \( x \) onto the vector \(w \) is given by \( y_i =w^\top x_i \). 

The objective of the FD is to find \(w \) that maximizes the SNR

\[
J(w) = \frac{(\mu_1 - \mu_2)^2}{s_1^2 + s_2^2},
\]

where \( \mu_1 =w^\top m_1 \) and \( \mu_2 =w^\top m_2 \) are the means of the projected points for classes \( C_1 \) and \( C_2 \), respectively, with \( m_1 = \frac{1}{n_1} \sum_{x_i \in C_1} x_i \) and \( m_2 = \frac{1}{n_2} \sum_{x_i \in C_2} x_i \) being the mean vectors of the classes in the original feature space. The terms \( s_1^2 \) and \( s_2^2 \) represent the variances of the projected points for classes \( C_1 \) and \( C_2 \), respectively.

To express \( J(w) \) in terms of the original data, we can define the within-class scatter matrix \( S_w \) and the between-class scatter matrix \( S_b \) as

\[
S_w = \sum_{x_i \in C_1} (x_i - m_1)(x_i - m_1)^\top + \sum_{x_i \in C_2} (x_i - m_2)(x_i - m_2)^\top,
\]

\[
S_b = (m_1 - m_2)(m_1 - m_2)^\top.
\]

Then, we have that the Fisher criterion can then be rewritten as

\[
J(w) = \frac{w^\top S_bw}{w^\top S_ww}.
\]

Then to maximize \( J(w) \), we differentiate it with respect to \(w \) and set the derivative to zero. This leads to the generalized eigenvalue problem given by 

\[
S_bw = \lambda S_ww.
\]

The solution \(w \) that maximizes \( J(w) \) is the eigenvector corresponding to the largest eigenvalue of the matrix \( S_w^{-1} S_b \).

The FD is closely related to Linear Discriminant Analysis (LDA) in the following sense. In LDA, the assumption is that the data from each class is normally distributed with a common covariance matrix \( \Sigma \). The LDA decoding rule assigns a new observation \( x \) to the class \( C_1 \) or \( C_2 \) based on which class’s linear discriminant function is maximized with

\[
\delta_k(x) = x^\top \Sigma^{-1} m_k - \frac{1}{2} m_k^\top \Sigma^{-1} m_k + \log \pi_k,
\]

where \( \pi_k \) is the prior probability of class \( C_k \), and \( k \in \{1, 2\} \). The LDA solution involves projecting the data onto a lower-dimensional space using the same criterion as the Fisher Linear Discriminant, making the FD effectively equivalent to LDA when the data follows a Gaussian distribution with equal covariance matrices across classes. This is why we track the Fisher Discriminant in our synthetic model: because we see mean and median LDA posterior probability on the correct class increase during overtraining, suggesting that SNR as measured by Fisher Discriminant is also increasing. 

The connection between FD and margin maximization can be understood in simpler, more qualitative terms: by examining the concept of separating hyperplanes in the context of binary classification. In margin-based methods such as SVMs, the goal is to find a hyperplane that maximizes the margin, which is the distance between the hyperplane and the nearest points from each class (support vectors). The FD also seeks to maximize the separation between classes, but it does so by maximizing the ratio of the between-class variance to the within-class variance. It can be thought of as a ``soft margin" of sorts. 

If the data is linearly separable, the direction \(w \) found by FD will tend to align with the direction that maximizes the margin between classes, although FD does not explicitly enforce the maximum margin criterion as SVM does. However, in many cases, the Fisher criterion leads to a solution that approximates the maximum margin direction, particularly when the distributions of the classes are approximately Gaussian with equal covariances, which is why in practice these two quantities often tend to vary together, as indeed they do both in our neural data and our synthetic model. Regardless, they measure different things.

\subsection{Representational Separation Ablation}\label{appdx: ablations}

\begin{figure}
    \centering
    \includegraphics[width=\linewidth]{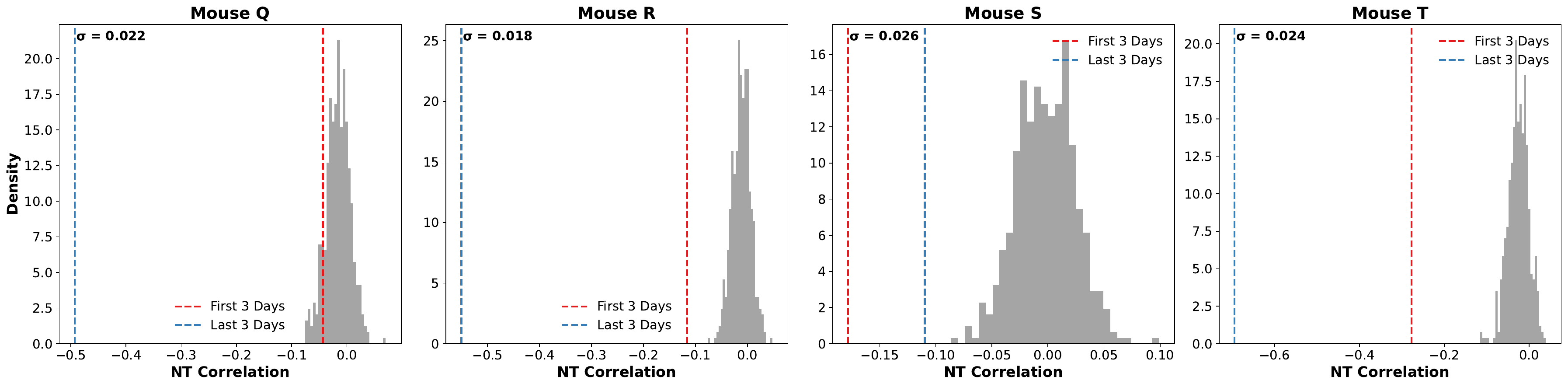}
    \caption{Cluster separation (anticorrelation between representations for classes) over 500 random firing rate matrices mean-matched with our z-scored firing rate matrix. $\sigma$ denotes standard deviation. More negative is better.}
    \label{fig:ablation}
\end{figure}

Since we z-score the firing rates, the diagonals of the correlatation matrices will be positive by definition as they take the form $||z||$. However, the expectation of the off-diagonals is zero for a set of random firing rates, with variance depending on dimension. If we are claiming there is significant separation in representations, we must check that the anticorrelation between the two firing rates is far from chance for our dimension $D \approx 15$. In Figure \ref{fig:ablation}, we plot the distribution of anticorrelations for a mean-matched firing rate matrix with firing rate vectors drawn from a Gaussian for each mouse, and draw a line to represent the anticorrelation (representational quality, since the task is to discriminate two classes) both before and after the overtraining period. We plot these in Figure \ref{fig:ablation}. 

We can see generally that 1) overtraining causes separation, in the sense that representations are more anticorrelated after overtraining compared to training, and that 2) even at the beginning of overtraining (red lines) the separation is nontrivial compared to chance. This makes sense because the mice can perform the task at ceiling, so we expect representations at the beginning of overtraining (after a week of overtraining) to be much better than chance. The caveats are that the data is noisy: for instance, Mouse S has anticorrelation slightly decrease compared to the beginning of overtraining. Such limitations of the data emphasize that our findings are \textit{suggestive} rather than conclusive, and we aim to inspire further experimental work and pose our findings as \textit{hypotheses} that this data suggest are \textit{plausible} rather than definitively validated. 

\subsection{Hard Margin Ablations}
\label{appdx:hard-margin}

We begin with a review of the Cross-Entropy vs Hinge loss definitions. Let $y \in\{0,1\}$ represent the true class label, and $\hat{y} \in[0,1]$ represent the predicted probability for the positive class (i.e., $P(y=1 \mid x)$ ). The cross-entropy loss $L_{\mathrm{CE}}$ for a single example is defined as:
$$
L_{\mathrm{CE}}(y, \hat{y})=-(y \log (\hat{y})+(1-y) \log (1-\hat{y})).
$$
In hinge loss, the labels $y$ are typically encoded as $y \in\{-1,+1\}$, and the model outputs a score $f(x) \in \mathbb{R}$, where the sign of $f(x)$ determines the class prediction. For simplicity, let's assume $f(x)$ is the output of a linear model, $f(x)=w \cdot x+b$.

\begin{figure}
    \centering
    \includegraphics[width=0.9\linewidth]{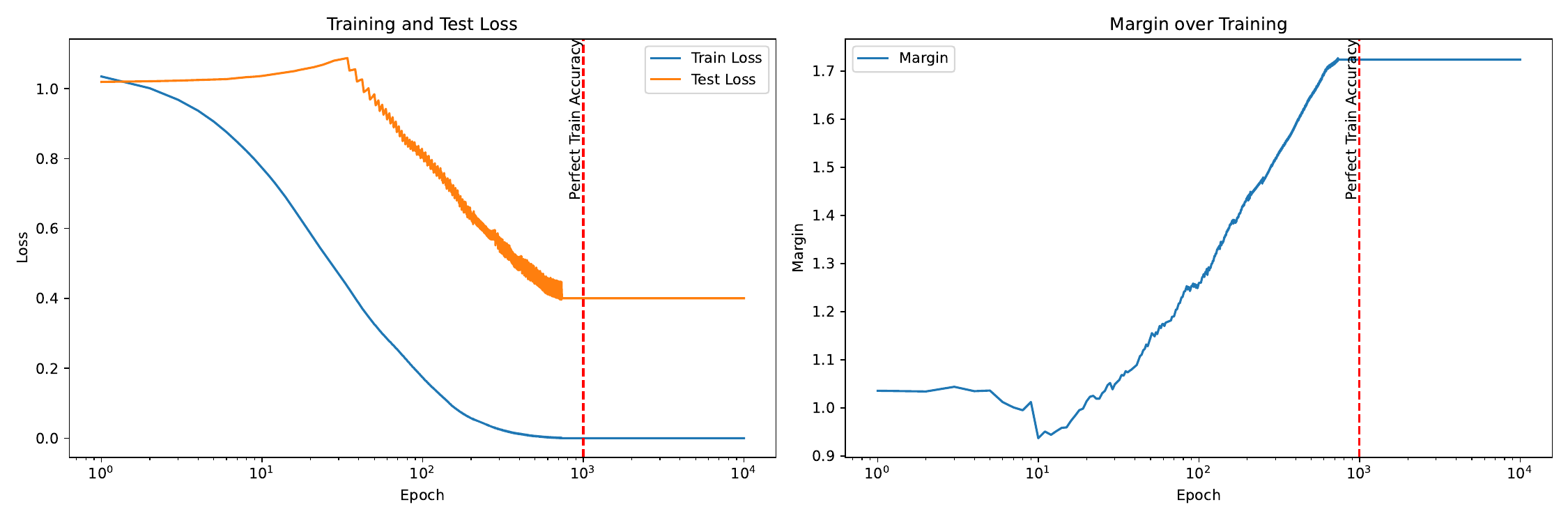}
    \includegraphics[width=0.95\linewidth]{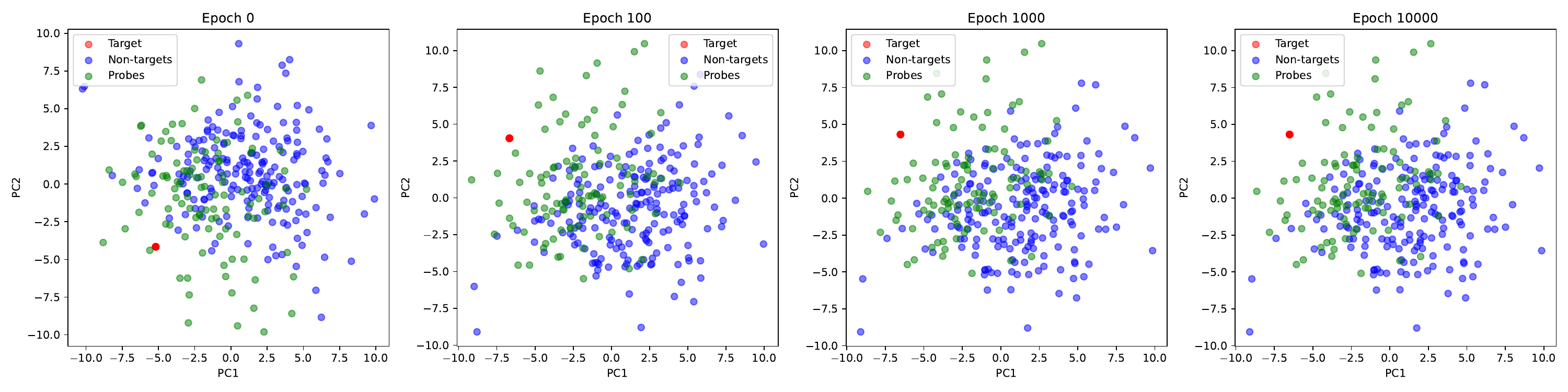}
    \caption{Loss function ablation for synthetic model that traces asymptotic margin maximization induced by cross-entropy as the cause for late-time decrease in test loss/late-time separation of representations. ``Late-time" refers to when training loss plateaus at a low value.}
    \label{fig:mlp_hinge}
\end{figure}

The hinge loss $L_{\text {hinge }}$ for a single example is defined as:
$$
L_{\text {hinge }}(y, f(x))=\max (0,C-y \cdot f(x)),
$$
where $C$ is the hard margin parameter to be specified, which we take as $C=1$. The key point is that in a Hinge objective, loss vanishes after the learned classifier classifies each training example with margin at least $C$, whereas loss does not vanish in this discontinuous way for the Cross-Entropy objective, driving continued late-time learning. We can see in Figure \ref{fig:mlp_hinge}that test loss stops improving when train loss does, and this is coincedent with an abrupt ceasation in the increase in margin, in contrst to the continued margin maximization driven by Cross-Entropy in the main text. 

\subsection{Biologically Plausible Model}
\label{appdx:biop}

We now repeat the above experimental setup of our synthetic model with an architecture that takes piriform cortex more seriously \citep{stern2018transformation, krishnamurthy2022disorder, mathis2016reading}.

Our biologically plausible model consists of three main processing layers plus an output layer, with additional feedback connections to model recurrent dynamics. The first layer serves as a sensory input layer, analogous to inputs from the olfactory bulb. It projects the input to a higher-dimensional space, using ReLU activation followed by dropout with probability 0.2 to maintain sparse coding, a key feature of sensory representations in the piriform cortex. The second layer acts as a dense associative layer, modeling the extensive recurrent connectivity between pyramidal cells. This layer maintains the expanded dimensionality from layer one and includes a parallel feedback pathway. The feedback is implemented through a separate linear transformation followed by ReLU activation, and the result is added back to the main pathway via a skip connection. Both the main and feedback pathways use the same dimensionality to preserve the associative nature of this layer. The third layer implements modulatory control, inspired by inhibitory interneuron circuits. This layer reduces dimensionality compared to layer two and uses LeakyReLU activation followed by a higher dropout rate (0.3) to model inhibitory effects. The reduction in dimensionality reflects the modulatory rather than representational role of this layer. The output layer performs a final linear transformation to a single scalar value (since we're training on binary classification). We train without weight decay and learning rate 1e-2. We also add Gaussian noise to the forward pass to simulate stochastic activity in sensory cortex, hence why there are multiple target odor representations.

\begin{figure}
    \centering
    \includegraphics[width=0.9\linewidth]{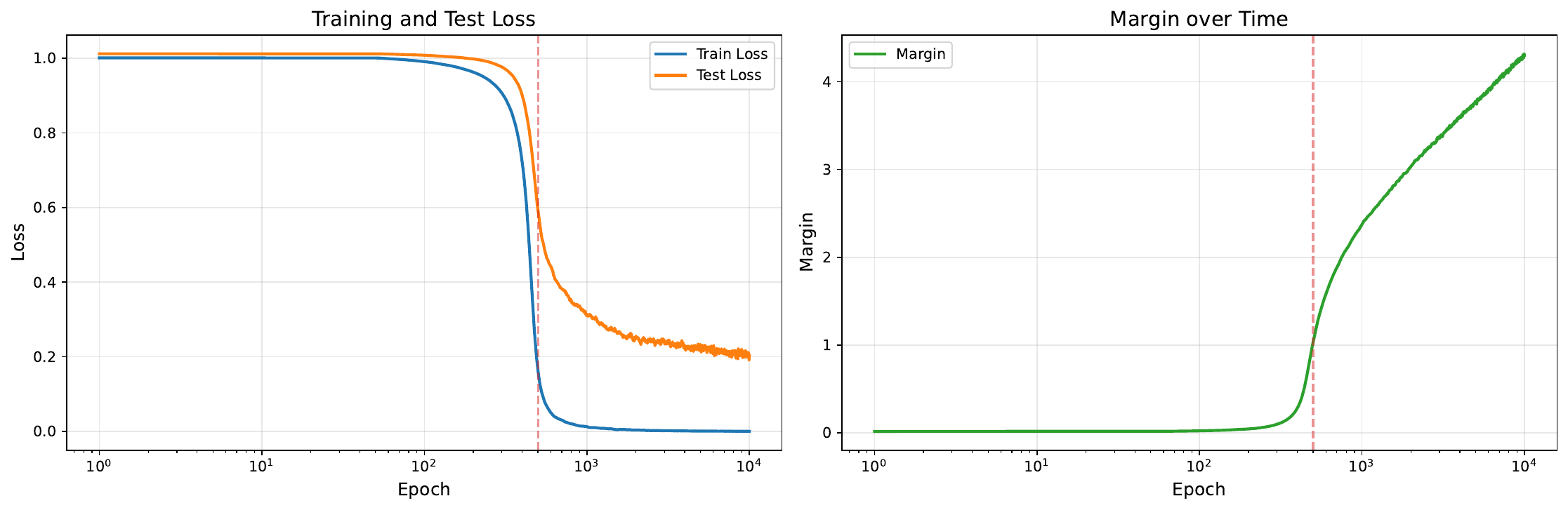}
    \includegraphics[width=0.95\linewidth]{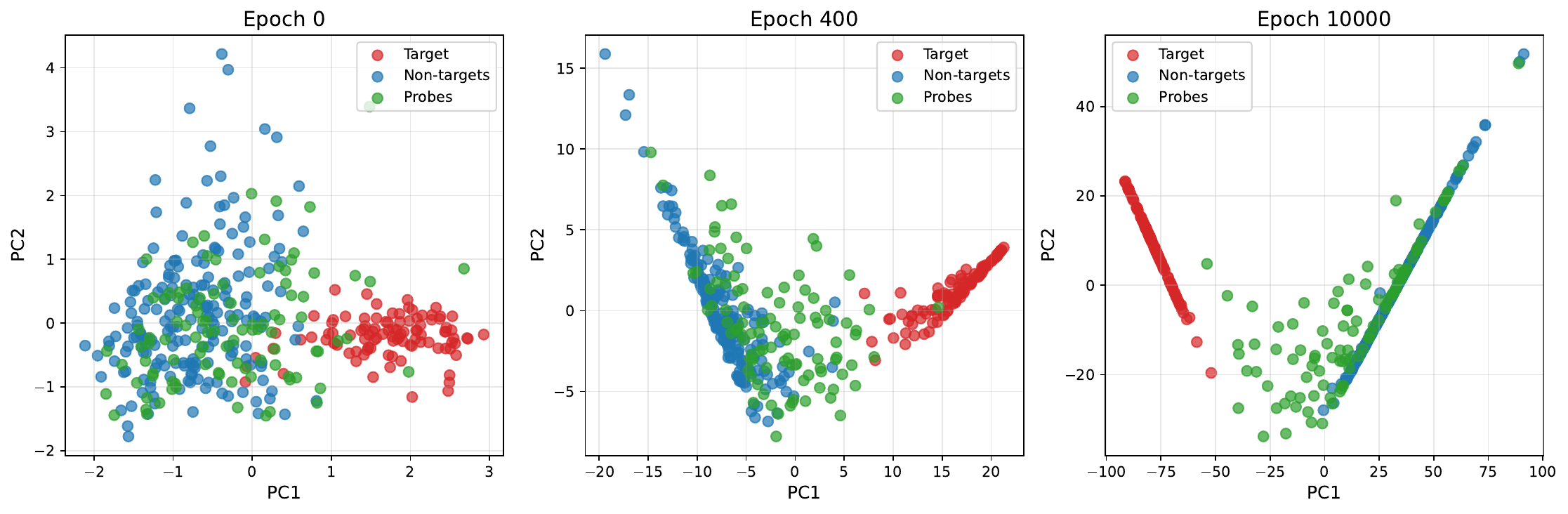}
    \caption{Architecture ablation for biologically plausible model of piriform cortex trained on the same binary classification task to separate target and nontarget odors. We see loss on held out probe odors continues to decrease after low train loss is achieved, since cross-entropy loss is once again used as the objective.}
    \label{fig:mlp_hinge_bio}
\end{figure}

Results are shown in Figure \ref{fig:mlp_hinge_bio}. The persistence of the phenomenology identified in the main text suggests that the objective function and task structure, rather than architectural details, are the key here.

\end{document}